\documentclass{ieeeaccess}
\usepackage{cite}
\usepackage{amsmath,amssymb,amsfonts}
\usepackage{graphicx}
\usepackage{textcomp}
\usepackage{caption}
\usepackage{acronym} 
\usepackage[tight,footnotesize]{subfigure}
\DeclareMathOperator*{\argmax}{arg\,max}

\usepackage{tabularx}
\usepackage{lscape}
\usepackage{mathtools}
\usepackage{booktabs}
\usepackage{algpseudocode}
\usepackage{algorithm}
\newcommand{\kk}[1]{}
\usepackage[none]{hyphenat}
\def\BibTeX{{\rm B\kern-.05em{\sc i\kern-.025em b}\kern-.08em
    T\kern-.1667em\lower.7ex\hbox{E}\kern-.125emX}}
\DeclareUnicodeCharacter{2212}{-}

\begin{document}
\history{Date of publication xxxx 00, 0000, date of current version xxxx 00, 0000.}
\doi{}

\title{SAAM: Stealthy Adversarial Attack on Monocular Depth Estimation}
\author{\uppercase{Amira Guesmi}\authorrefmark{1}, 
\uppercase{Muhammad Abdullah Hanif\authorrefmark{1}, Bassem Ouni \authorrefmark{2}, and Muhammad Shafique}.\authorrefmark{1}
}
\address[1]{eBrain Lab, Division of Engineering, New York University (NYU) Abu Dhabi, UAE}
\address[2]{AI and Digital Science Research Center, Technology Innovation Institute (TII), Abu Dhabi, UAE}


\markboth
{Guesmi \headeretal: SAAM: Stealthy Adversarial Attack on Monocular Depth Estimation}
{Guesmi \headeretal: SAAM: Stealthy Adversarial Attack on Monocular Depth Estimation}

\corresp{Corresponding author: Amira Guesmi (e-mail: ag9321@nyu.edu).}

\begin{abstract}
Monocular depth estimation (MDE) is an important task in scene understanding, and significant improvements in its performance have been witnessed with the utilization of convolutional neural networks (CNNs). These models can now be deployed on edge devices, thanks to advancements in CNN optimization, enabling effective depth estimation in safety-critical and security-sensitive systems like robots, rovers, drones, and autonomous cars. However, CNNs used for MDE are susceptible to adversarial attacks, which can be exploited for malicious purposes by generating plausible images containing carefully crafted perturbations that distort the model's output.
To assess the vulnerability of CNN-based depth prediction methods, recent studies have attempted to design adversarial patches specifically targeting MDE. However, these methods have not been powerful enough to fully deceive the vision system in a systemically threatening manner. Their impact is less effective, misleading the depth prediction of only certain parts within the overlapping region of the input image by using conspicuous and eye-catching patterns.
In this paper, we investigate the vulnerability of MDE to adversarial patches. We propose a novel \underline{S}tealthy \underline{A}dversarial \underline{A}ttacks on \underline{M}DE (SAAM) that compromises MDE by either corrupting the estimated distance or causing an object to seamlessly blend into its surroundings. Our experiments demonstrate that the designed stealthy patch successfully causes a CNN to misestimate the depth of objects. 
In fact, our proposed adversarial patch achieves a significant 60\% depth error with 99\% ratio of the affected region. Importantly, despite its adversarial nature, the patch maintains a naturalistic appearance, making it inconspicuous to human observers. 
We believe that this work sheds light on the threat of adversarial attacks in the context of MDE on edge devices. We hope it raises awareness within the community about the potential real-life harm of such attacks and encourages further research into developing more robust and adaptive defense mechanisms.
\end{abstract}


\titlepgskip=-15pt

\maketitle
\section{Introduction}
\label{sec:introduction}
\begin{figure*}[!t]
    \centering
    \includegraphics[width=0.7\textwidth]{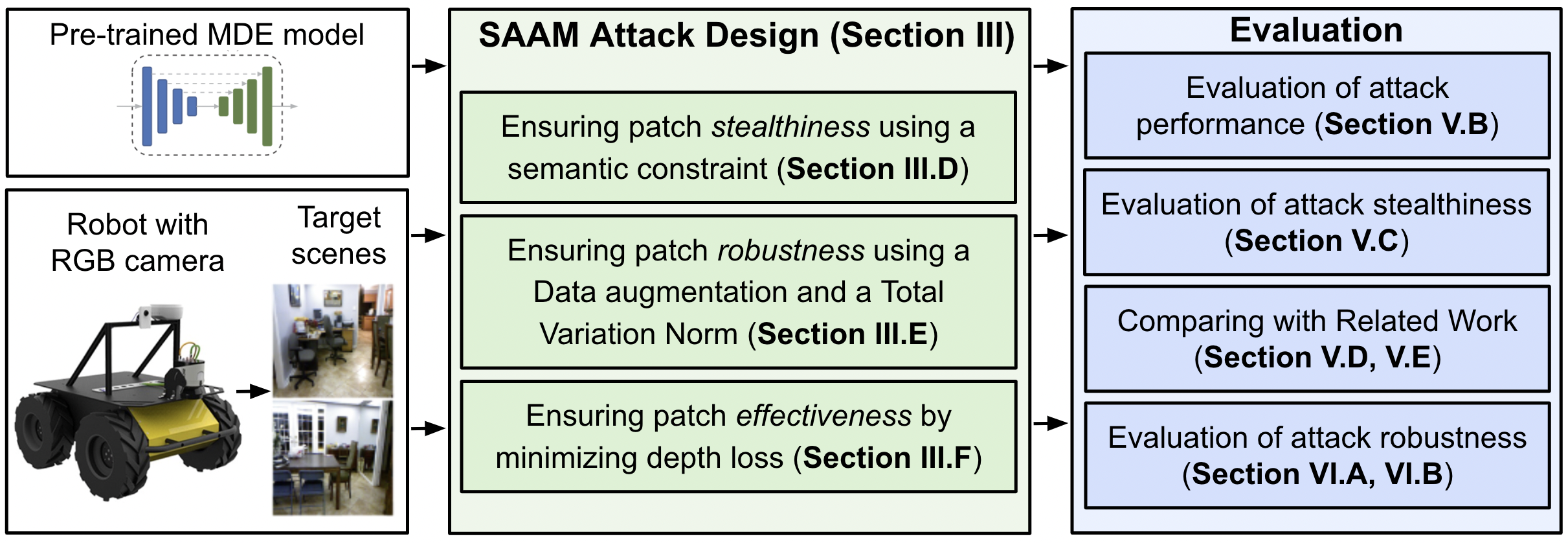}
    \caption{Overview of our Novel Contributions.}
    \label{contribution}
\end{figure*}
\PARstart{M}{onocular} depth estimation (MDE) is increasingly being utilized in a wide array of real-world applications, ranging from autonomous driving to robotics. Its main purpose is to acquire depth data, enabling a deeper and more comprehensive comprehension of the surrounding scene. MDE plays a critical and indispensable role in various tasks, including obstacle avoidance \cite{obstacles}, object detection \cite{autonomous2}, visual SLAM \cite{robotics1, slam}, visual re-localization \cite{relocalization}, and numerous others.

Several methods for estimating depth rely on sensors like RGB-D cameras, radar, LIDAR, or ultrasound to collect the depth information directly from a scene. However, these have significant shortcomings. In fact, ultrasound devices suffer from inherently imprecise measurements, LIDAR and radar produce sparse information, and RGB-D cameras have a narrow measuring range. The aforementioned devices are extremely large and power hungry for small-sized systems, especially those that must adhere to rigorous real-world design constraints. In contrast, RGB cameras are lightweight and less expensive. More importantly, they can offer more detailed environmental information.
Several tasks can now be accomplished totally using MDE's measurement and attain competitive performance thanks to MDE's rapid developments.
These remarkable improvements can be attributed to the successful integration of deep neural networks, which have significantly enhanced MDE's capabilities.

However, the increasing reliance on deep neural networks also brings attention to their vulnerability to adversarial attacks. As shown in various studies, these networks can be susceptible to manipulations that intentionally deceive their predictions. Hence, it becomes crucial to prioritize the security of MDE models to ensure their reliability and trustworthiness in practical applications. Safeguarding MDE systems against adversarial attacks is vital for preserving their integrity and preventing potential misinterpretations and misjudgments in real-world scenarios.

Patch-based adversarial attacks \cite{guesmi2023dap, guesmi2023aparate, guesmi2023advart} are a type of adversarial attack in computer vision, where carefully crafted perturbations are applied to specific patches or regions of an input image to deceive a deep learning model. The goal of patch-based adversarial attacks is to cause the model to misclassify the entire image or produce incorrect predictions for targeted regions. In contrast to traditional global adversarial attacks that perturb the entire input image, patch-based attacks are more localized, focusing on specific regions of interest. 

Patch-based adversarial attacks have implications in various applications, including object detection, image segmentation, and scene understanding. They demonstrate the vulnerability of deep learning models to localized adversarial perturbations and highlight the importance of developing robust defense mechanisms to protect against such attacks. 


Only a limited number of studies have explored the realm of patch-based adversarial attacks on depth estimation. This particular direction of research remains relatively unexplored compared to other adversarial attack methods in computer vision.
Previous work for patch-based adversarial attacks on MDE \cite{Yamanaka_Access, Cheng_ECCV, guesmi2023aparate} aiming at tricking the perception module of an autonomous vehicle. Their effectiveness is limited, as they only mislead the depth prediction of specific parts within the overlapping region between the input image and the patch, usually utilizing conspicuous and eye-catching patterns.
In this work, we investigate stealthy adversarial patches that can either fully conceal a particular object or trick the target methods into estimating the depth of that object incorrectly\/ at a target depth.

\begin{figure*}[!t]
    \centering
    \includegraphics[width=0.9\textwidth]{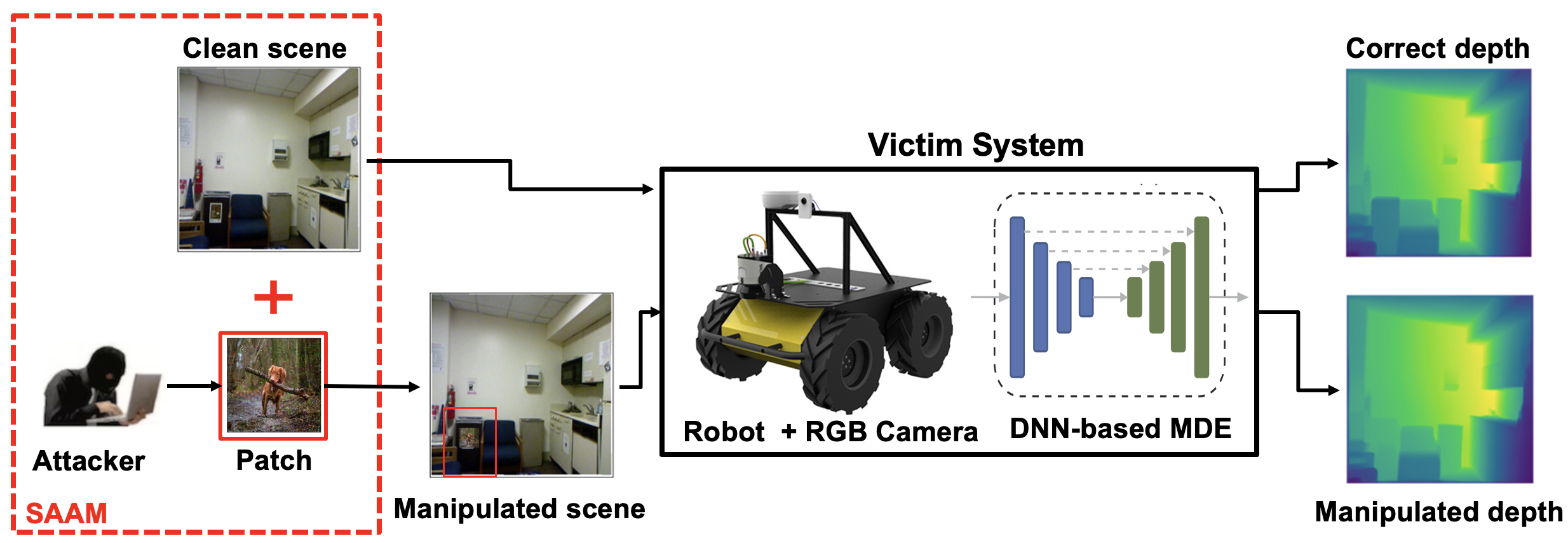}
    \caption{Overview of our proposed attack SAAM. In the first row, the depth estimation for objects in a clean scene is accurate and correctly predicted by the MDE-based system. However, in the second row, when the objects are manipulated using our adversarial patch, the depth estimation for these objects becomes incorrect and cannot be accurately estimated by the system.}
    \label{saam}
\end{figure*}

This paper introduces a technique for deceiving a CNN-based monocular depth estimation system by leveraging naturalistic adversarial patches (See Figure \ref{saam}). These patches are strategically designed to manipulate the system's predictions, resulting in the generation of false distance estimates. The proposed approach allows the adversarial patch to seamlessly blend into its surroundings while for example resembling a painting on a wall or a poster, and can be applied to conceal specific objects or areas of interest effectively.
In fact, we set out to achieve two key goals: depth manipulation and object concealment (i.e., object-background blending). Through our proposed techniques, we can intentionally alter the perceived depth of specific objects in a scene, leading to inaccurate depth estimations. Additionally, we have developed a method to completely conceal certain objects from the depth estimation process, making them effectively invisible to the system. Moreover, our approach enables selected objects to seamlessly blend with the background, creating a visual effect where they appear to be part of the scenery, thus reducing their conspicuousness and detection. An overview of our novel contributions is shown in Figure \ref{contribution}.

\noindent
\textbf{In summary, the contributions of this work are: }

\begin{itemize}
    \item We present a novel patch-based adversarial attack that targets DNN-based monocular depth estimation.
    \item Our framework generates a stealthy adversarial patch (SAAM) that can seamlessly blend into its surroundings (e.g., resembling a painting on a wall or a poster).
    \item SAAM has the ability to withstand diverse transformations and adapt to different scenarios. It demonstrated robustness against a range of deformations, including rotation, perspective change, and lighting variation. Additionally, the patch can be placed at arbitrary locations within the scene, even under occlusion.
    \item Our proposed adversarial patch extends its applicability to multiple use cases such as navigation tasks, obstacle detection, localization, etc. 
\end{itemize}

With a patch size as small as 0.7\% of the input image, we achieve an impressive 60\% depth estimation error. Moreover, the adversarial patch nearly covers the entire target region, with an almost 100\% ratio of the affected region. These findings highlight the effectiveness and potency of our attack in disrupting the depth estimation process.

The structure of the remaining article is organized as follows. Section \ref{sec:related} provides a comprehensive overview of related work in the field of patch-based adversarial attacks on monocular depth estimation. 
In Section \ref{sec:proposed}, we present our proposed methodology to generate the adversarial patch. This section outlines the step-by-step process of crafting the patch and explains the techniques employed to achieve effective depth manipulation and visual realism. Section \ref{sec:setup} details the experimental setup used to evaluate the performance of the proposed adversarial attack. In Section \ref{sec:experiment}, we delve into the evaluation of the proposed attack. We present the results obtained from various metrics, such as depth error, affected region ratio, and SSIM, to assess the attack's potency and visual similarity of the generated patch. In Section \ref{sec:discussion}, we thoroughly discuss the findings and implications of our experiments. We analyze the strengths and limitations of the proposed attack and interpret the results in the context of real-world applications. Section \ref{sec:conclusion} provides a succinct summary and conclusion of our study. 

\begin{table}
\caption{Notations used in this paper.}
\label{table}
\begin{tabular}{cl}
\hline
Notation & Description \\
\hline
$I$& Clean Image \\
$I^*$& Adversarial Image \\
$N$ & Natural Image \\
$M_p$ & Mask matrix to constrain the size, the shape and placement  \\
$T_\theta$  & Ensemble of transformations\\
$P_\delta$   & The adversarial patch\\
$F$  & DNN-based MDE model \\
$d_{clean}$    & Clean depth estimation\\
$d_{adv}$    & Adversarial depth estimation\\
$\left\|.\right\|_p $  & $L_p$ norm\\
$\delta$ & Adversarial noise\\
$\epsilon$ & Maximum allowable magnitude of perturbation\\
$L_{tv}$ & Total Variation loss\\
$L_{depth}$   & Depth loss\\
$L_{total}$& Total loss\\
$c$    & Target depth\\
$\alpha, \beta$   & Loss weights\\ 
$\beta_1, \beta_2$   & Optimizer parameters\\ 
$l_r$    &  Learning rate\\ 
$E_d$    &  Depth estimation error\\
$R_a$    & Ratio of affected region\\
\hline
\end{tabular}
\label{tab1}
\end{table}

\section{ Background and Related Work}
\label{sec:related}
\subsection{Monocular Depth Estimation}
Monocular depth estimation is a prominent computer vision task that involves predicting the depth information of a scene from a single 2D image \cite{monodepth2, godard2019digging}. It holds fundamental significance in computer vision and finds applications in a wide range of fields, including robotics, augmented reality, and autonomous vehicles. The main objective of monocular depth estimation is to infer the 3D structure of the scene solely from a single 2D image. Unlike stereo depth estimation, which relies on multiple images captured from slightly different viewpoints, monocular depth estimation's practicality lies in its ability to work with just one image, making it well-suited for various real-world scenarios.
\begin{figure*}[!t]
    \centering
    \includegraphics[width=0.7\textwidth]{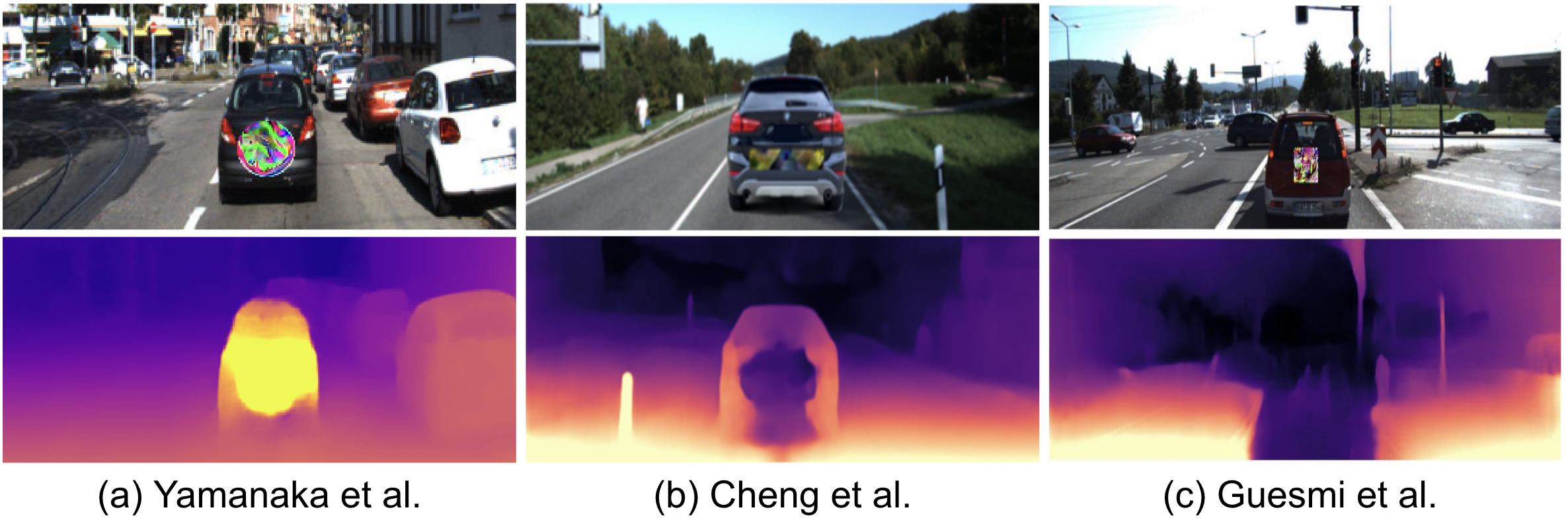}
    \caption{State of the art patch-based adversarial attacks on MDE.}
    \label{related}
\end{figure*}

 \begin{table*}[!ht]
  \centering
  \caption{Comparison of attack methods. M - Manipulate estimation; H – Hiding objects. A – Anywhere in the frame; O – On the target object(s)}
  \label{versus}
  \begin{tabular}{|l|c|c|c|c|c|}
    \hline
    \textbf{Method} & \textbf{Attack goal} &\textbf{Attacker’s Knowledge} &\textbf{Stealthy}  &\textbf{Placement} &\textbf{Setting}\\
    \hline
      \textbf{Adversarial patch \cite{Yamanaka_Access}}  &  M    & White-box & $\times$    & A  &  Outdoor  \\
      \hline
     \textbf{AOP \cite{Cheng_ECCV}}  &  M    & White-box &  $\checkmark $    &  O   &  Outdoor   \\
      \hline
     \textbf{APARATE \cite{guesmi2023aparate}}  &  M, H    & White-box &   $\times$     &  O  &  Outdoor  \\
     \hline
     \textbf{SAAM (ours)}  &   M, H   & White-box &  $\checkmark $  &   O, A & Indoor     \\
  \hline
\end{tabular}
\end{table*}
\subsection{Adversarial Examples}
An attacker possesses the flexibility to modify the input image of a victim model at the pixel level. These attacks inherently assume that the attacker has control over the DNN's input system, such as a camera. The first adversarial example, proposed in \cite{Szegedy}, involved adding small imperceptible noise to steer the prediction of the input image towards an incorrect class. Numerous attacks have been developed, advancing the algorithms for generating adversarial examples \cite{CW, fgsm, BA, localsearch, HSJ}. In the face of the persistent challenges posed by adversarial attacks, researchers have diligently worked to establish resilient defenses, consistently pushing the boundaries of innovation \cite{10130393, chattopadhyay2023oddr, chattopadhyay2023defensivedr}.

Although there have been efforts to enhance the power of these attacks by crafting real-time attacks \cite{moosavidezfooli2017universal, room} that are generated on the fly, a more robust and realistic threat model would consider the scenario where the attacker has exclusive control over the system's external environment or external objects, rather than its internal sensors and data pipelines. In the following sections, we will explore some state-of-the-art patch-based attacks on object detectors.

\subsection{Patch-based adversarial attacks on MDE}
Patch-based attacks are a specific form of adversarial perturbation that focuses on modifying localized patches or regions within an image with the intention of deceiving machine learning models. 
These perturbations are carefully designed and printed on physical surfaces, such as images or objects, to exploit the vulnerabilities of computer vision systems. They aim to introduce imperceptible alterations that can lead to misclassifications or incorrect interpretations by the targeted models. This method involves substituting a section of the targeted image with an image patch to impede the performance of DNN-based models. 
The adversarial patch is prevalent because it is simple to use and can usually be printed out with ease.

Yamanaka et al. \cite{Yamanaka_Access} was the first to propose a method for generating printable adversarial patches for corrupting MDE-based systems. However, the patches generated in their approach had eye-catching patterns, making them easily noticeable. Cheng et al. \cite{Cheng_ECCV} focused on addressing the issue of stealthiness in the generated patch, aiming to ensure that the patch is inconspicuous and does not draw attention. However, a drawback of the generated patch is that it is object-specific, meaning that a separate patch needs to be trained for each target object. Additionally, the patch had a limited affected region and was trained for a specific setting with a fixed distance between the object and the camera, making it ineffective for other distances. Guesmi et al. \cite{guesmi2023aparate} proposed an adaptive adversarial patch optimized to be shape and scale-aware, and its impact adapts to the target object instead of being limited to the immediate neighborhood. Although this patch was effective but didn't consider the appearance and the stealthiness of the generated patterns. Figure \ref{related} illustrates the state-of-the-art patch-based adversarial attacks on Monocular Depth Estimation (MDE). Meanwhile, Table \ref{versus} provides a comprehensive comparison of these attack methods, considering various aspects: attack goal, attacker’s Knowledge, attack stealthiness, the placement of the patch, and the setting specifying whether the attack is designed for indoor or outdoor scenes.

\section{Proposed Approach}
\label{sec:proposed}

\subsection{Problem formulation}

In the context of Monocular Depth Estimation, when presented with a benign image $I$, the objective of the adversarial attack is to compromise the depth estimation process by employing a maliciously designed adversarial patch $P_{\delta}$. This patch is strategically crafted to introduce stealthy perturbations into the original image, transforming it into an adversarial example denoted as $I^*$.
Technically, the adversarial example with generated patch can be formulated as:

\begin{equation}
    I^* = (1 - M_P) \odot I + M_P \odot T_{\theta}(P_{\delta})
\end{equation}
$\odot$ is the component-wise multiplication, $P_{\delta}$ is the adversarial patch, $T_{\theta}$ is the ensemble of patch transformations, and $M_P$ is a mask matrix to constrain the shape, the size and pasting position of the patch, where the value of the pasting area is 1 and 0 elsewhere.

The adversarial depth, i.e., the output of the victim model when taking as input the adversarial example is:
\begin{equation}
   d_{adv} = F((1 - M_P) \odot I + M_P \odot T_{\theta}(P_{\delta}))
\end{equation}
The problem of generating an adversarial example can be formulated as a constrained optimization \ref{eq:adv}, given an original input image $I$ and a DNN-based MDE model $ F(.) $:
\begin{equation}
\label{eq:adv}
\begin{aligned}
    \min_{\delta} \left\|\delta\right\|_{\infty}  \\
     s.t.~~ F((1 - M_P) \odot I + M_P \odot T_{\theta}(P_{\delta})) \neq d_{clean}\\ 
\end{aligned}
\end{equation}

The objective is to find a minimal adversarial noise, denoted as $\delta$ used to form the adversarial patch $P_{\delta}$, which, when applied at any arbitrary placement in the scene, selectively undermines the underlying MDE model $F(.)$ by causing the objects to appear farther or closer than they really are. It is important to note that a closed-form solution cannot be obtained for this optimization problem due to the non-convex nature of the DNN-based MDE model $F(.)$. Therefore, Equation \ref{eq:adv} can be reformulated as follows to enable numerical approximation of the problem using empirical techniques:

\begin{equation}
\label{eq:formulation}
    \argmax_{P} \sum_{I \in \mathcal{U}} loss(F((1 - M_P) \odot I + M_P \odot P), d_{clean}) 
\end{equation}
Where $loss$ is a predefined loss function and $\mathcal{U} \subset U$ is the attacker’s training dataset.
We can use existing optimization techniques (e.g., Adam \cite{adam}) to solve this problem. In each iteration of the training the optimizer updates the adversarial patch $P$.

\subsection{Adversarial Patch Evaluation}
In the evaluation of physical adversarial attacks, three key aspects are commonly considered: effectiveness, robustness under real-world conditions, and stealthiness to human observers.
\begin{itemize}
    \item \textbf{Effectiveness: }
    Attack effectiveness is a critical aspect to consider when assessing the impact of physical attacks. These attacks have demonstrated their effectiveness in significantly degrading the performance of the targeted task, thereby compromising the reliability and accuracy of the victim system. Adversarial manipulations in the physical space can cause misclassifications, incorrect predictions, or erroneous decisions, leading to potentially severe consequences.
    \item \textbf{Robustness:} 
    Attack robustness is a key factor in evaluating the resilience of physical attacks in dynamic environments. Maintaining attack ability despite variations in the environment is crucial for the sustained effectiveness of adversarial manipulations. One aspect of robustness is being able to maintain attack efficacy across different scenes. This means that the perturbations should remain effective and capable of deceiving the system, regardless of changes in lighting conditions, backgrounds, or other scene-specific factors.  
    \item \textbf{Stealthiness:} 
    Attack stealthiness is a critical characteristic that determines the effectiveness of physical attacks. To be successful, these attacks should ideally go unnoticed by both the observer and the victim, remaining imperceptible to human eyes. The ability to maintain stealthiness ensures that the adversary can carry out their attack without raising suspicion or triggering any defensive mechanisms.
\end{itemize}

\subsection{Overview of our Methodology}
\begin{figure*}[!t]
    \centering
    \includegraphics[width=0.8\textwidth]{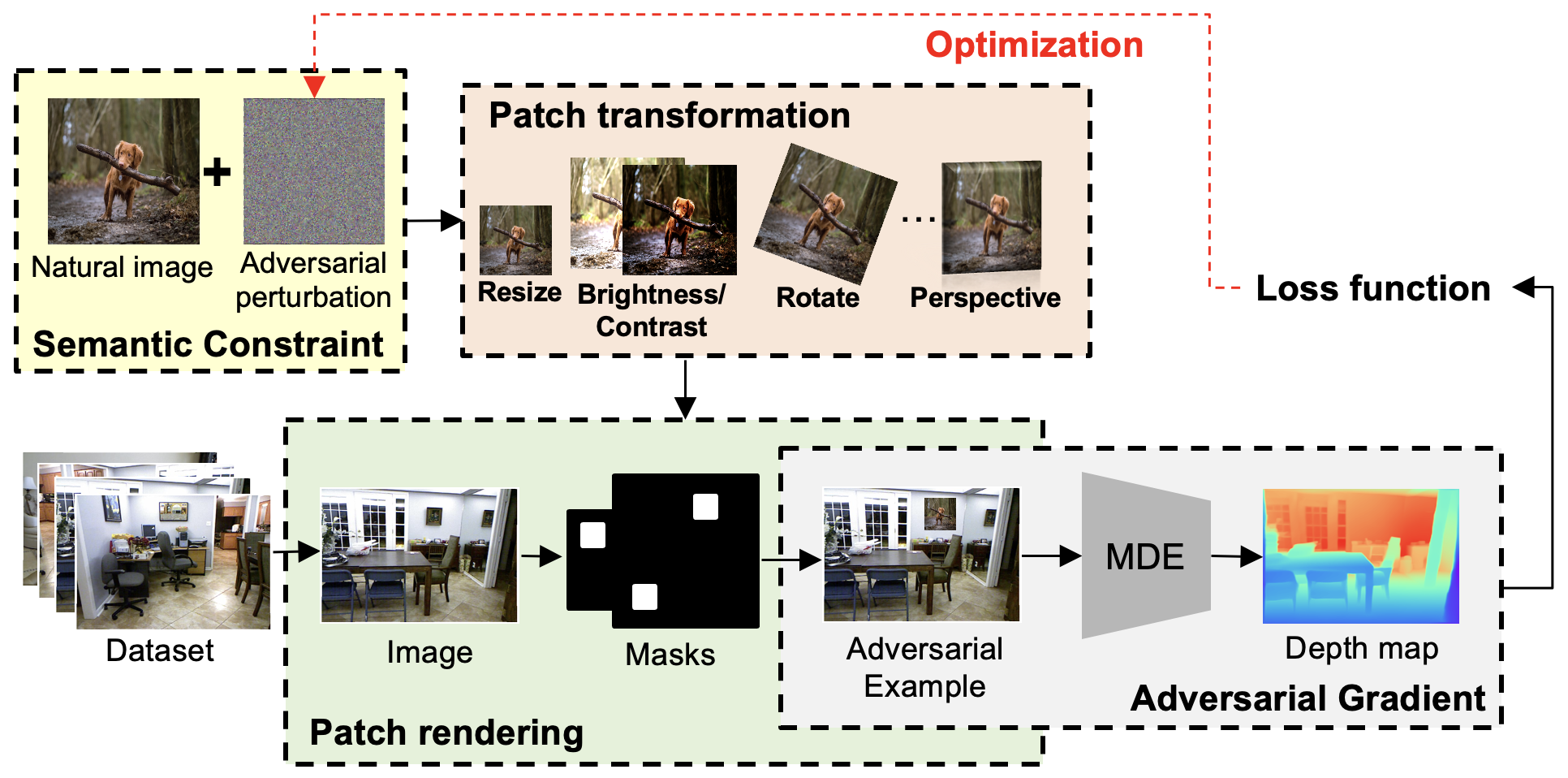}
    \caption{Overview of the stealthy adversarial patch (SAAM) framework used to generate patches for monocular depth estimation. This framework leverages a semantic constraint to ensure the stealthiness of the generates adversarial patch and a data augmentation technique which takes potential transformation in the real world into account during the optimization.}
    \label{fig:approach}
\end{figure*}
In our approach, illustrated by Figure \ref{fig:approach}, we start by introducing randomly initialized noise to a selected natural image. This noise serves as the basis for our adversarial patch generation. To enhance the robustness of the patch and mimic real-world scenarios, we apply various data augmentation techniques, such as resizing, adjusting brightness and contrast, rotation, and perspective changes. These transformations help us create diverse versions of the patch that could occur in practical situations.
Next, we proceed with the patch rendering process. We utilize generated masks, representing different and random placements of the patch within the scene, to superimpose the patch on the clean image of the scene. This composite image, which now contains the adversarial patch, is then fed into the deep neural network-based monocular depth estimation (MDE) model to generate the corresponding depth map.
To optimize the adversarial perturbation and ensure its effectiveness, we compute the loss function based on the discrepancy between the depth map generated with the adversarial patch and the clean depth values. Our goal is to maximize this loss function to achieve the highest impact in manipulating the depth prediction.
However, it is crucial to maintain the semantic meaning of the patch and retain visual similarity to the original natural image. To address this, we enforce a semantic constraint on the adversarial perturbation. We project the noise onto the surface of an $L_p$ norm-ball with a predefined radius $\epsilon$. This constraint helps to preserve the meaningful appearance of the patch while keeping it within a reasonable perturbation range.
By following these steps, we can generate effective adversarial patches capable of concealing objects or altering their perceived depth in the scene. Our approach ensures that the patches are practically applicable, maintain visual realism, and successfully deceive the deep neural network-based MDE model.

\subsection{Patch Stealthiness: Semantic Constraint}
Taking inspiration from the imperceptibility constraint commonly employed in Lp-norm based adversarial perturbations, we incorporate a projection function (Equation \ref{semantic_constraint}) to ensure that the generated adversarial patterns maintain visual similarity to natural images throughout the optimization process. By enforcing this constraint, we achieve high-quality semantic patterns that closely resemble a predefined natural image, for example, a painting on a wall. Empirical results demonstrate the effectiveness of optimizing with this constraint, as it facilitates the creation of visually convincing adversarial patterns that seamlessly blend into their surroundings.
\begin{equation}
    P_{\delta} = N + \delta
\end{equation}
Where $N$ is a chosen natural images to ensure the generated camouflage patterns are semantically meaningful.

\begin{equation}
    \delta^t = Proj_{\infty}(\delta^{(t-1)} + \Delta \delta, N, \epsilon)
    \label{semantic_constraint}
\end{equation}

where $\delta^t$ and $\Delta \delta$ denote the adversarial pattern and its updated vector at iteration t, respectively. $Proj_{\infty}$ projects generated pattern onto the surface of $L_{\infty}$ norm-balls with radius $\epsilon$ and centered at N. Here we choose N as natural images to ensure the generated camouflage patterns are semantically meaningful. For our experiments we set $\epsilon=0.3$.

\subsection{Patch Robustness}
\subsubsection{Data Augmentation}
In order to effectively deceive CNN-based monocular depth estimation models in real-world scenarios, we incorporate the considerations of physical world conditions during the optimization process of adversarial patches. Real-world scenarios often involve various conditions, including changing lighting, different viewpoints, natural noise, and more. To simulate such dynamic factors, we apply several physical transformations. These transformations encompass various aspects, such as adding noise, random rotation, varying scales, random brightness and contrast adjustments, and more. These physical transformation operations are encapsulated within the patch transformer.

The geometric transformations performed include randomly scaling the patch $[0.25, 1.25]$. Additionally, random rotations ($\pm20^\circ$) are applied to the patch $P_\delta$. This simulates uncertainties in patch placement, size, and distance with respect to the camera. Color space transformations are conducted by introducing random noise ($\pm0.1$) to pixel intensity values, applying random contrast adjustments within the range of $[0.8, 1.2]$, and implementing random brightness adjustments ($\pm0.1$). Furthermore, we perform patch cropping and perspective change.
This process leads to the formation of the resulting patch $T_\theta (P_\delta)$, which is then forward propagated through the monocular depth estimator.

By accounting for these physical transformations and incorporating them into the optimization process, we aim to create adversarial patches that can successfully deceive CNN-based monocular depth estimation models in real-world scenarios. In typical real-world applications, the object detection model scales the adversarial sample along with the image to a square shape, which can reduce the effectiveness of the patch. To address this limitation, SAAM directly scales the adversarial patch along with the targeted images. This scaling operation introduces additional deformation to the patch, making it more challenging to optimize.
\begin{figure*}
    \centering
    \includegraphics[width=0.7\textwidth]{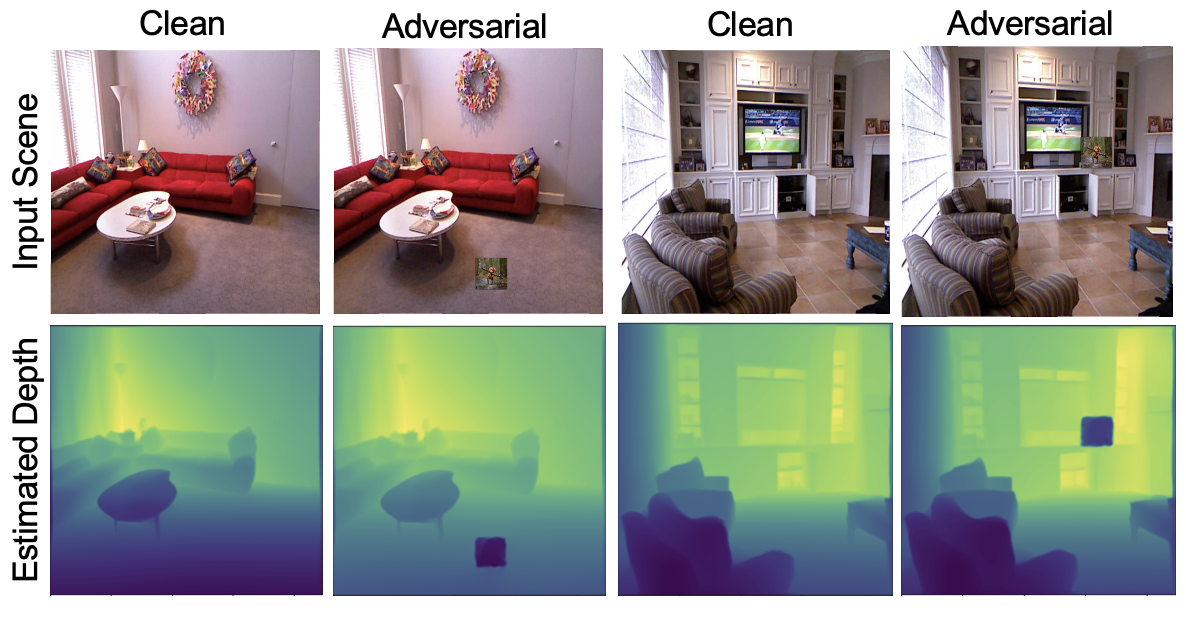}
    \caption{Estimated depth for different patch locations (patch scale: 1\%): the patch is designed to create an illusion of the target region appearing closer.}
    \label{fig:depth_closer}
\end{figure*}
\subsubsection{Total Variation Norm (TV loss)}
The characteristics of natural images include smooth and consistent patches with gradual color changes within each patch \cite{mahendran2015understanding}. Therefore, 
To increase the plausibility of physical attacks, smooth and consistent perturbations are preferred. Additionally, extreme differences between adjacent pixels in the perturbation may not be accurately captured by cameras due to sampling noise. This means that non-smooth perturbations may not be physically realizable \cite{Sharif2016FaceRecognitionAttacks}. To address these issues, the total variation (TV) \cite{mahendran2015understanding} loss is introduced to maintain the smoothness of the perturbation.
For a perturbation $P$, TV loss is defined as:

It is defined as:
\begin{equation}
    L_{tv} = \sum_{i,j} \sqrt{(P_{i+1,j} - P_{i,j})^2 + (P_{i,j+1} - P_{i,j})^2}
\end{equation}

where the subindices $i$ and $j$ refer to the pixel coordinate of the patch $P$.


\subsection{Adversarial Patch Generation}
We iteratively perform gradient updates on the adversarial patch $(P_{\delta})$ in the pixel space in a way that optimizes our objective function defined as follows:

\begin{equation}
    L_{total} = \alpha L_{depth} + \beta L_{tv} 
\end{equation}

$L_{depth}$ is the adversarial depth loss.
\begin{equation}
    d_{adv} = F((1 - M_P) \odot I + M_P \odot T_{\theta}(P_{\delta}))
    \label{pre}
\end{equation}
The adversarial losses are defined as the distance between the estimated adversarial depth (Eq. \ref{pre}) and the estimated clean depth or the target depth and calculated as follows:\\

\noindent For \textit{un-targeted} attacks:
\begin{equation}
    L_{depth} = -(\left | d_{clean} - d_{adv} \right | \odot M_P )
    \label{untargeted}
\end{equation}
\noindent For \textit{targeted} attacks:
\begin{equation}
    L_{depth} = d_{clean} \odot (M_P \times c)
    \label{targeted}
\end{equation}
Where $c$ is the target depth.

$L_{tv}$ is the total variation loss on the generated image to encourage smoothness.

$\alpha$, and $\beta$  are hyper-parameters used to scale the losses. For our experiments we set $\alpha = 1$ and $\beta = 0.5$.
We optimize the total loss using Adam \cite{adam} optimizer. We try to minimize the object function $L_{total}$ and optimize the adversarial patch. We freeze all weights and biases in the depth estimator and only update the pixel values of the adversarial patch. The patch is randomly initialized. 

The optimization process involves iteratively updating the patch to maximize the adversarial objective function. This is achieved through gradient-based optimization, where the gradient of the objective function with respect to the patch is calculated.
\begin{equation}
    P_{new} = P_{old} + \lambda \cdot \nabla_{P} L_{total}(P_{old},I)
\end{equation}

where $\lambda$ is the learning rate. To maintain consistency with the original range of the clean input image, we apply a clamping operation, restricting the values of the patch between 0 and 1. This ensures that the adversarial patch remains within the valid range and aligns with the characteristics of the unperturbed input image.

\section{Experimental Setup}
\label{sec:setup}
In order to assess the effectiveness of our proposed attack, we analyzed vulnerabilities of two DNN-based MDE models;  The self-supervised depth prediction models are chosen based on their practicality and open source codes.
\begin{figure*}
    \centering
    \includegraphics[width=0.9\textwidth]{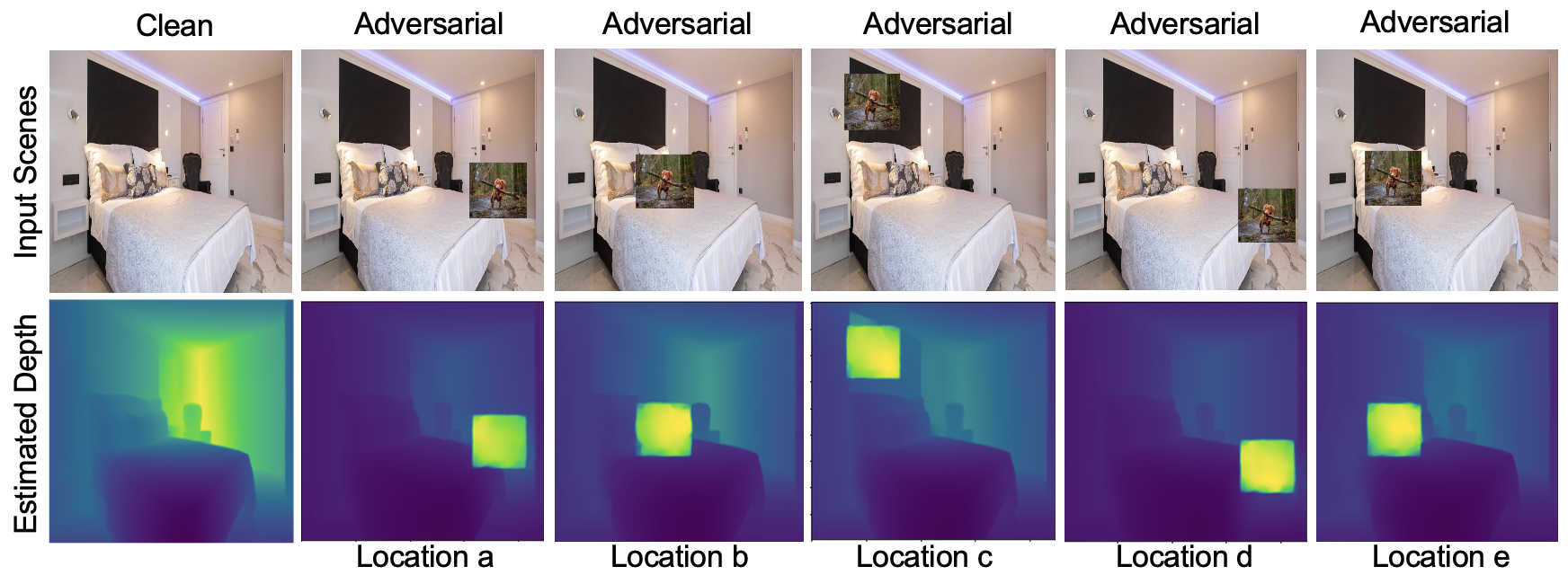}
    \caption{Estimated depth for different patch locations (patch scale: 5\%): the patch is designed to create an illusion of the target region appearing farther.}
    \label{fig:depth_farther}
\end{figure*}

\noindent\textbf{Networks:}
DiverseDepth \cite{Wei2021CVPR} trained on depth prediction on multiple data sources including high-quality LiDAR sensor data \cite{Zamir_2018_CVPR}, and low-quality web stereo data \cite{Katrin2019,Chaoyang2019, xian2020structure}. 
The model has a backbone ResNet-50 and ResNet-101. 
The DiverseDepth model, is a cutting-edge deep learning architecture specifically designed for monocular depth estimation. It addresses the challenge of handling diverse real-world scenarios by introducing an affine-invariant representation of the depth map, ensuring accurate predictions regardless of camera orientation or viewpoint changes. Leveraging multi-task learning, the model combines depth estimation with tasks like normal estimation and instance segmentation, enhancing overall scene understanding and boosting depth prediction robustness. The model benefits from extensive data augmentation, simulating diverse and realistic scenarios through geometric transformations to enrich the training dataset. Trained on a large-scale dataset encompassing various scenes, camera poses, and lighting conditions, the DiverseDepth model excels in predicting depths accurately under real-world variations. Additionally, it estimates depth uncertainty, crucial for assessing the reliability of predictions in different image regions. As a result, DiverseDepth achieves state-of-the-art performance on benchmark datasets, demonstrating its significant contribution to advancing monocular depth estimation.

Monodepth2 \cite{monodepth2} is based on the general U-Net architecture, i.e. an encoder-decoder network, enabling the representation of both deep abstract features as well as local information. The ResNet18 \cite{resnet} pretrained on ImageNet \cite{imagenet} was used as the encoder and the decoder was based on several convolution and upsampling layers with skip connections used to decode the output back to the input resolution.
Monodepth2 builds upon the original Monodepth model, significantly improving depth estimation performance. Notably, Monodepth2 adopts an unsupervised learning approach, utilizing monocular video sequences for training without the need for ground truth depth annotations. The model employs a geometry-based loss to enforce consistency in predicted depth and ego-motion across consecutive frames, encouraging accurate depth estimation aligned with the scene's geometry. Its encoder-decoder architecture captures multi-scale features and employs skip connections for enhanced depth prediction. Furthermore, Monodepth2 offers the capability of estimating monocular ego-motion, making it valuable for comprehensive scene understanding and visual odometry tasks. The model's unsupervised nature and ability to predict both depth and ego-motion have made it a significant advancement in monocular depth estimation, yielding competitive results on benchmark datasets while reducing the reliance on costly ground truth annotations.

GLPdepth \cite{glpdepth} involves a hierarchical transformer encoder to capture global context and a lightweight, efficient decoder for local connectivity. They employ a selective feature fusion module to connect multi-scale local features and the global decoding stream, allowing the network to produce more accurate depth maps with fine details. Additionally, the proposed decoder outperforms previous models with lower computational complexity. They enhance depth-specific augmentation techniques based on an important depth estimation observation, achieving state-of-the-art results on the NYUv2 dataset. Their approach demonstrates better generalization and robustness compared to other models.

MIMDepth \cite{mimdepth} is an approach that leverages masked image modeling (MIM) for self-supervised monocular depth estimation. While MIM has traditionally been used for pre-training to learn generalizable features, authors proposed its adaptation for the direct training of monocular depth estimation. Their experiments revealed that MIMDepth exhibits greater robustness to various challenges, including noise, blur, weather conditions, digital artifacts, occlusions, as well as both untargeted and targeted adversarial attacks.

\noindent\textbf{Datasets:} The patch was trained on indoor scenes from NYUv2 dataset \cite{nyuv2}. 
The NYUv2 dataset is a widely used benchmark dataset for depth estimation and 3D scene understanding in computer vision. It was introduced by Silberman et al. in 2012 and is an extension of the original NYU Depth dataset. The NYUv2 dataset provides RGB-D data, consisting of RGB images and corresponding depth maps, captured from a variety of indoor scenes. The dataset contains images and depth maps from a diverse set of indoor scenes, captured with Microsoft Kinect cameras. The scenes include various rooms, objects, and furniture arrangements. Each RGB-D data sample in the dataset includes a high-resolution RGB image (640x480 pixels). The RGB images capture the color information of the indoor scenes. The dataset provides aligned depth maps for each RGB image, obtained from the Kinect depth sensor. The depth maps contain per-pixel depth information, allowing researchers to perform monocular depth estimation and other 3D scene understanding tasks. The NYUv2 dataset comprises a significant amount of data, with over 1449 RGB-D samples. This large-scale nature makes it suitable for training and evaluating deep learning models. The indoor scenes in the dataset cover a wide range of challenging scenarios, including occlusions, cluttered environments, and varying lighting conditions.


We conducted our experiments on a TeslaV100 GPU. The models were implemented using Python version 3.10.12 and PyTorch \cite{NEURIPS2019_9015} version 2.1.0.
In our optimization process, we employed the Adam optimizer \cite{adam}, with a learning rate ($l_r$) set at 0.001, along with $\beta1 = 0.9$ and $\beta2 = 0.999$ for momentum parameters. Our optimization procedure extended across 200 epochs. The details of various attack hyperparameters can be found in Table \ref{hyper}, and a comprehensive list of applied transformations is available in Table \ref{transformations}.

Same as in \cite{Yamanaka_Access}, the patch resolution was configured at $256\times256$ pixels, yet its apparent size within the input image was changed by the patch transformation block.

\begin{table}[!htp]
  \centering
  \caption{Attack hyper-parameters.}
  \label{hyper}
  \begin{tabular}{|l|c|}
    \hline
    \textbf{Parameters} &  \textbf{Values} \\
     \hline
    $\epsilon$ & 0.03\\
    $c$    & $20~ m$\\
    ($\alpha, \beta$)   & ($1, 0.5$)\\
    ($\beta_1, \beta_2$)   &  ($0.9, 0.999$) \\
    $l_r$    &  $0.001$\\
    $epochs$    &  $200$\\
    $batch$    &  $8$\\    
\hline
\end{tabular}
\end{table}

\begin{table}[!htp]
  \centering
  \caption{Transformation distribution.}
  \label{transformations}
  \begin{tabular}{|l|c|c|}
    \hline
    \textbf{Transformations} &  \textbf{Parameters} & \textbf{Remark} \\
    \hline
    Random Noise & $\pm0.1$ & Noise   \\
    Rotation    & $\pm20^\circ$ & Camera Simulation   \\
    Brightness  & $\pm0.1$ & Illumination   \\
    Contrast    & $[0.8, 1.2]$ & Camera Parameters   \\
    Cropping    & −0.7 $\sim$ 1.0 & Photograph/Occlude Simulation   \\
    Affine & $0.7$ & Perspective/Deformed Transforms  \\
    Scale    & $[0.25, 1.25]$ & Distance/Resize \\
  \hline
\end{tabular}
\end{table}

\section{Experiments}
\label{sec:experiment}
\begin{figure*}
    \centering
    \includegraphics[width=0.5\textwidth]{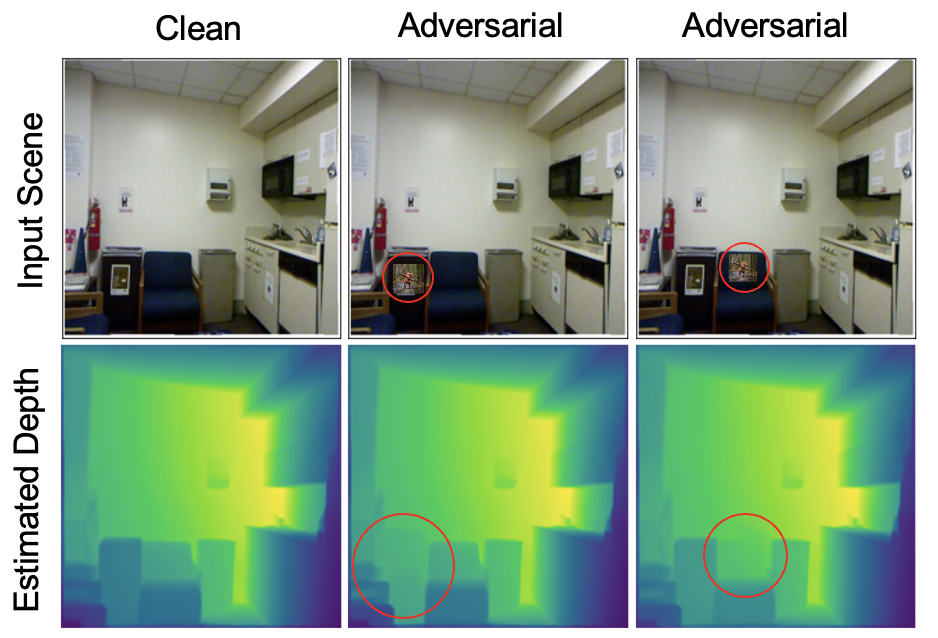}
    \caption{Effectiveness of SAAM in concealing objects: Placing the patch on target objects changes its predicted depth in a way that the object is blended with the background.}
    \label{fig:conceal}
\end{figure*}
To validate the effectiveness of our proposed method, we conducted a thorough analysis of the vulnerabilities exhibited by an MDE model trained on indoor scene data. Our approach involved generating adversarial examples to evaluate the model's robustness to potential attacks.
By crafting adversarial examples using our proposed technique, we were able to manipulate the input data in subtle yet strategic ways, aiming to deceive the MDE model during its depth estimation process. These adversarial examples were carefully designed to conceal specific objects, alter their perceived depths, or induce other misinterpretations, challenging the model's ability to accurately understand the scene.

We systematically evaluated the performance of the MDE model by feeding it these adversarial examples and comparing the resulting depth estimations with the ground truth data. Through comprehensive experimentation and analysis, we gained insights into the model's vulnerabilities and potential weaknesses, highlighting the need for robustness against adversarial attacks.
\subsection{Evaluation Metrics}

To assess the efficacy of our proposed adversarial attack, we rely on two widely used metrics employed in previous works \cite{Yamanaka_Access, guesmi2023aparate}: the mean depth estimation error ($E_d$) and the ratio of the affected region ($R_a$).

For calculating $E_d$, we consider the depth prediction of the adversarial object and compare it with the depth prediction of the benign object, using the latter as the ground truth. This metric quantifies the extent of the attack's effectiveness in altering the perceived depth of the affected region. A higher value of $E_d$ indicates a more successful attack, as it signifies a larger divergence between the predicted depth and the ground truth. The $R_a$ metric, on the other hand, measures the ratio of the affected region, i.e., the proportion of the scene where the depth estimation has been altered by the adversarial patch. A higher value of $R_a$ indicates a larger portion of the scene being affected by the attack, highlighting the patch's ability to conceal objects or modify their depth over a significant area.

By utilizing these metrics, we quantitatively evaluate the performance of our proposed adversarial attack, providing valuable insights into its effectiveness and impact on the depth estimation process. A higher mean depth estimation error and a larger affected region ratio indicate a more potent attack, reinforcing the significance of our findings and the need for robust defenses against such adversarial attacks on depth-based computer vision systems.

\textit{The depth estimation error ($E_{d}$)} is defined as follow:
\begin{equation}
    E_{d} = \frac{\sum_{i,j}(\arrowvert d_{clean} - d_{adv} \arrowvert \odot M_P) }{\sum_{i,j}{M_P}}
\end{equation}

$R_{a}$ measures the ratio of pixels that their depth value has changed above a certain threshold with respect to the number of pixels in $M_P$. Any change in pixel's depth value above $0.1$, that pixel is considered as affected. $\textbf{I(x)}$ is the indicator function that evaluates to 1 only when the condition $x$ is true.

\textit{The ratio of affected region ($R_{a}$)} is defined as follow:
\begin{equation}
    R_{a} = \frac{\sum_{i,j}\textbf{I}((\arrowvert d_{clean} - d_{adv} \arrowvert \odot M_P) > 0.1) }{\sum_{i,j}{M_P}}
\end{equation}
\subsection{SAAM Effectiveness} 
\subsubsection{In corrupting the depth of objects}
In the initial experiment, we evaluated the impact of adversarial patch attacks on the performance of the victim depth estimation model. We examined how the presence of the adversarial patch affected the accuracy and reliability of the depth predictions produced by the model. By comparing the results obtained with and without the adversarial patch, we were able to assess the effectiveness and potential vulnerabilities of the victim depth estimation model under adversarial conditions. 
In un-targeted attacks, the objective of the attacker is to corrupt the depth estimation of objects in the scene without setting a specific target depth. In our experiments, as depicted in Figure \ref{fig:depth_closer}, we demonstrate the effectiveness of our proposed adversarial patch in achieving this goal. By crafting a patch that covers only 1\% of the input image, we were able to generate an adversarial patch capable of significantly disrupting the target regions depth estimation. The results presented in Table \ref{ed_untargeted} illustrate the effectiveness of our proposed untargeted adversarial attack. Remarkably, even with a patch size as small as 0.7\% of the input image, we were able to achieve a substantial 57\% depth error. 

Observations indicate that both MIMDepth and GLPDepth demonstrate increased resilience against our adversarial patch. However, it's worth noting that a substantial level of corruption remains, with depth estimation error rates of 47\% and 49\%, respectively, when considering a patch scale of 5\%.

\begin{table}[!htp]
  \centering
  \caption{Effect of un-targeted adversarial patch on victim models in terms of depth estimation error ($E_{d}$).}
  \label{ed_untargeted}
  \begin{tabular}{|c|c|c|c|c|}
    \hline
    \textbf{Scale} & \textbf{DiverseDepth}& \textbf{Monodepth2}  &  \textbf{MIMdepth } & \textbf{GLPdepth } \\
    \hline
      0.7\%       &  0.57  & 0.53 & 0.41    &  0.43  \\ 
      \hline
      1\%         & 0.58   & 0.56  &  0.42   &  0.44  \\
      \hline
      2\%         & 0.6   & 0.57   &  0.45   & 0.47 \\
      \hline
      5\%         & 0.62   &  0.59  & 0.47    & 0.49 \\
  \hline
\end{tabular}
\end{table}

The findings in Table \ref{ra_untargeted} indicate that the adversarial patch's impact on the depth estimation is extensive, with an almost 100\% ratio of the affected region. This means that the vast majority of the overlapped region from the scene is influenced by the patch, leading to significant distortions in the depth perception.

\begin{table}[!htp]
  \centering
  \caption{Effect of un-targeted adversarial patch on victim models in terms of ratio of affected region ($R_{a}$).}
  \label{ra_untargeted}
  \begin{tabular}{|c|c|c|c|c|}
    \hline
    \textbf{Scale} & \textbf{DiverseDepth}& \textbf{Monodepth2}&  \textbf{MIMdepth} & \textbf{GLPdepth}\\
    \hline
      0.7\%       & 0.98   & 0.98 &   0.97  &  0.97 \\  
      \hline
      1\%         & 0.99   &  0.98 &  0.97   &  0.98 \\  
      \hline
      2\%         & 0.99   & 0.99  & 0.98    & 0.98  \\   
      \hline
      5\%         & 0.99   &  0.99 &  0.98   & 0.99 \\   
  \hline
\end{tabular}
\end{table}

The ability to disrupt the scene depth without a specific target depth in mind highlights the far-reaching consequences of adversarial attacks and raises important concerns for applications relying on accurate depth estimation, such as autonomous vehicles, robotics, and augmented reality. 

In a targeted attack setting, we use the depth loss presented in equation \ref{targeted}, we set $c = 1$ to indicate that the attacker's goal is to alter the depth of the target region to the farthest point in the scene as illustrated in Figure \ref{fig:depth_farther}. By crafting a patch that covers 5\% of the input image, we were able to generate an adversarial patch capable of significantly disrupting the overall scene depth estimation. The patch strategically alters the depth perception of multiple objects in the scene, causing distortions and misinterpretations by the depth estimation model.

\begin{table}[!htp]
  \centering
  \caption{Effect of targeted adversarial patch on victim models in terms of depth estimation error ($E_{d}$).}
  \label{ed}
  \begin{tabular}{|c|c|c|c|c|}
    \hline
    \textbf{Scale} & \textbf{DiverseDepth}& \textbf{Monodepth2}&  \textbf{MIMdepth } & \textbf{GLPdepth}\\
    \hline
      0.7\%       &  0.48  & 0.49 &  0.37   & 0.4  \\ 
      \hline
      1\%         & 0.58   & 0.55 &  0.38   & 0.42   \\
      \hline
      2\%         & 0.56   & 0.54 &  0.41   & 0.44  \\
      \hline
      5\%         & 0.6   &  0.57 &  0.43   & 0.46 \\
  \hline
\end{tabular}
\end{table}

The same patch achieves 60\% depth error with 99\% affected region. These findings further emphasize the substantial disruption caused by the adversarial patch on the depth estimation process. The high depth error and near-complete coverage of the scene's affected region indicate a widespread and significant distortion of the perceived depth across multiple objects and regions within the image.

Attacking both MIMDepth and GLPDepth resulted in substantial corraption of the estimated depth. For instance, as shown in Tables \ref{ed} and \ref{ra} using a patch of a scale 1\% resulted in an $E_d$ of $0.38$ and $0.42$ and a $R_a$ of $0.96$ and $0.97$ for MIMDepth and GLPDepth, respectively.
\begin{table}[!htp]
  \centering
  \caption{Effect of targeted adversarial patch on victim models in terms of ratio of affected region ($R_{a}$).}
  \label{ra}
  \begin{tabular}{|c|c|c|c|c|}
    \hline
    \textbf{Scale} & \textbf{DiverseDepth}& \textbf{Monodepth2}&  \textbf{MIMdepth } & \textbf{GLPdepth}\\
    \hline
      0.7\%       & 0.97   & 0.97 & 0.96    & 0.96   \\  
      \hline
      1\%         & 0.99   &  0.98 &  0.96   &  0.97 \\  
      \hline
      2\%         & 0.98   & 0.98 & 0.97    &  0.98 \\   
      \hline
      5\%         & 0.99   &  0.99 &  0.98   & 0.98 \\   
  \hline
\end{tabular}
\end{table}

%
\begin{figure*}
    \centering
    \includegraphics[width=0.7\textwidth]{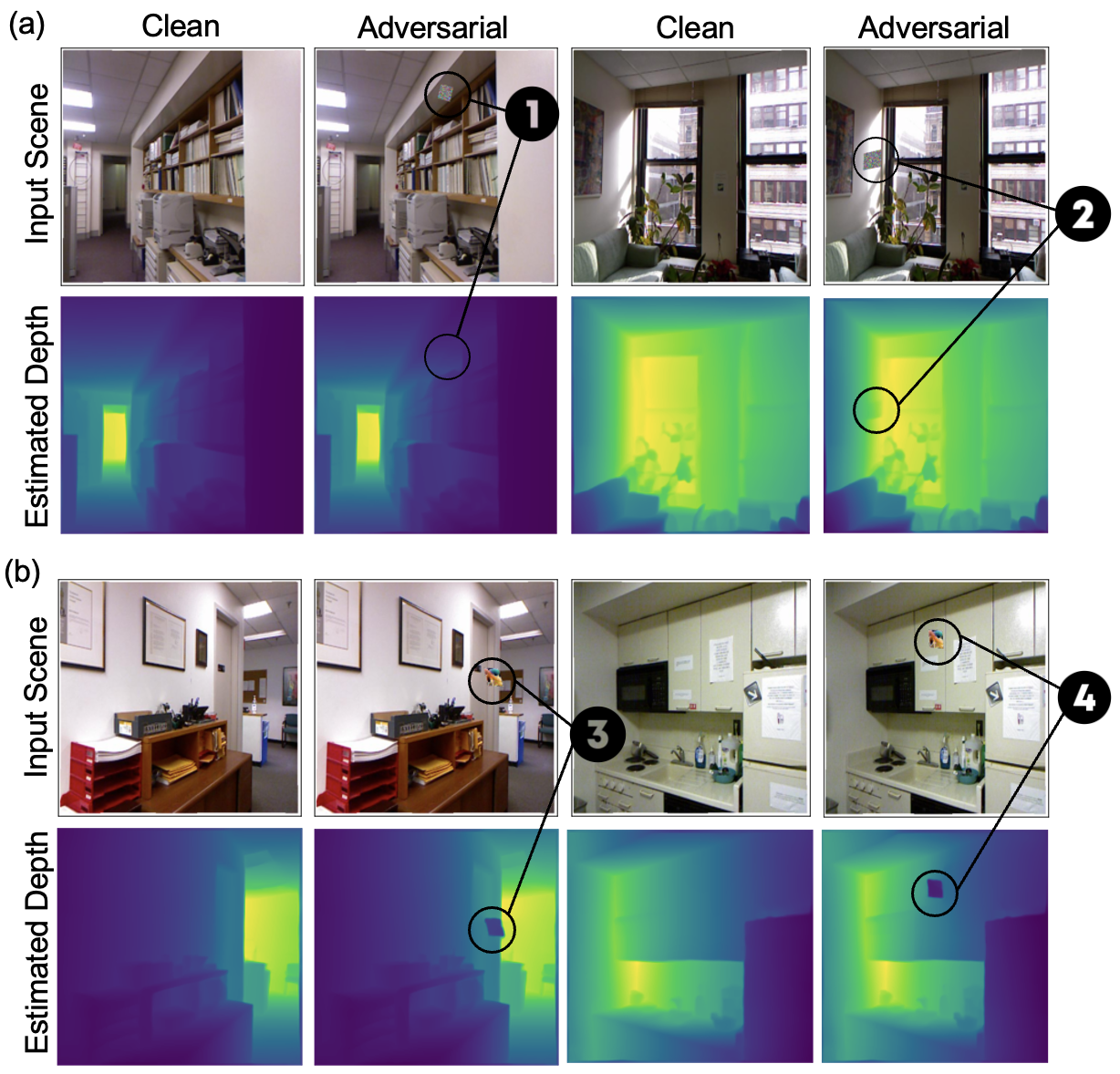}
    \caption{The comparison of the impact of (a) Random noise and (b) SAAM on the model performance under different transformations reveals distinct effects on the depth estimation model.}
    \label{fig:random}
\end{figure*}
\subsubsection{In concealing objects}
In this section, we assess the effectiveness of our attack in achieving the objective of concealing objects and making them blend with the background. By applying the patch strategically to specific objects within the scene, we analyze the resulting depth estimations and observe how effectively the objects remain concealed. We explore the capability of our adversarial attack to seamlessly blend targeted objects with the background. By manipulating the depth estimations, we aim to achieve a visual effect where the objects appear as natural components of the scene, effectively reducing their visibility and distinctiveness. We evaluate the blending performance through both quantitative metrics and visual assessments, comparing the modified scenes with their original counterparts to discern the extent of successful blending. As shown in Figure \ref{fig:conceal}, by setting the variable $c$ equal to the depth of the surroundings, our proposed adversarial patch successfully achieves the effect of making the target object blend seamlessly with its environment. 

\subsection{SAAM Stealthiness}

The Structural Similarity Index (SSIM) is a widely used metric to quantify the similarity between two images, assessing how close the generated patch is to the original benign target image. In our evaluation, as presented in Table \ref{ssim}, the computed SSIM value between the generated patch and the natural image is exceptionally high, with a score of 0.91 for a patch scale equal to 1\%.
Such a high SSIM score indicates that the adversarial patch is highly similar to the natural image, making it visually indistinguishable to human observers. 

 \begin{table}[!ht]
  \centering
  \caption{Structural Similarity Index (SSIM) between the benign target image and the generated patch for different patch scales.}
  \label{ssim}
  \begin{tabular}{|l|c|c|}
    \hline
    \textbf{Scale} & \textbf{1\%}  &\textbf{5\%} \\
    \hline
      \textbf{SSIM}  &   0.91   &   0.93     \\
  \hline
\end{tabular}
\end{table}

\begin{figure*}
    \centering
    \includegraphics[width=0.8\textwidth]{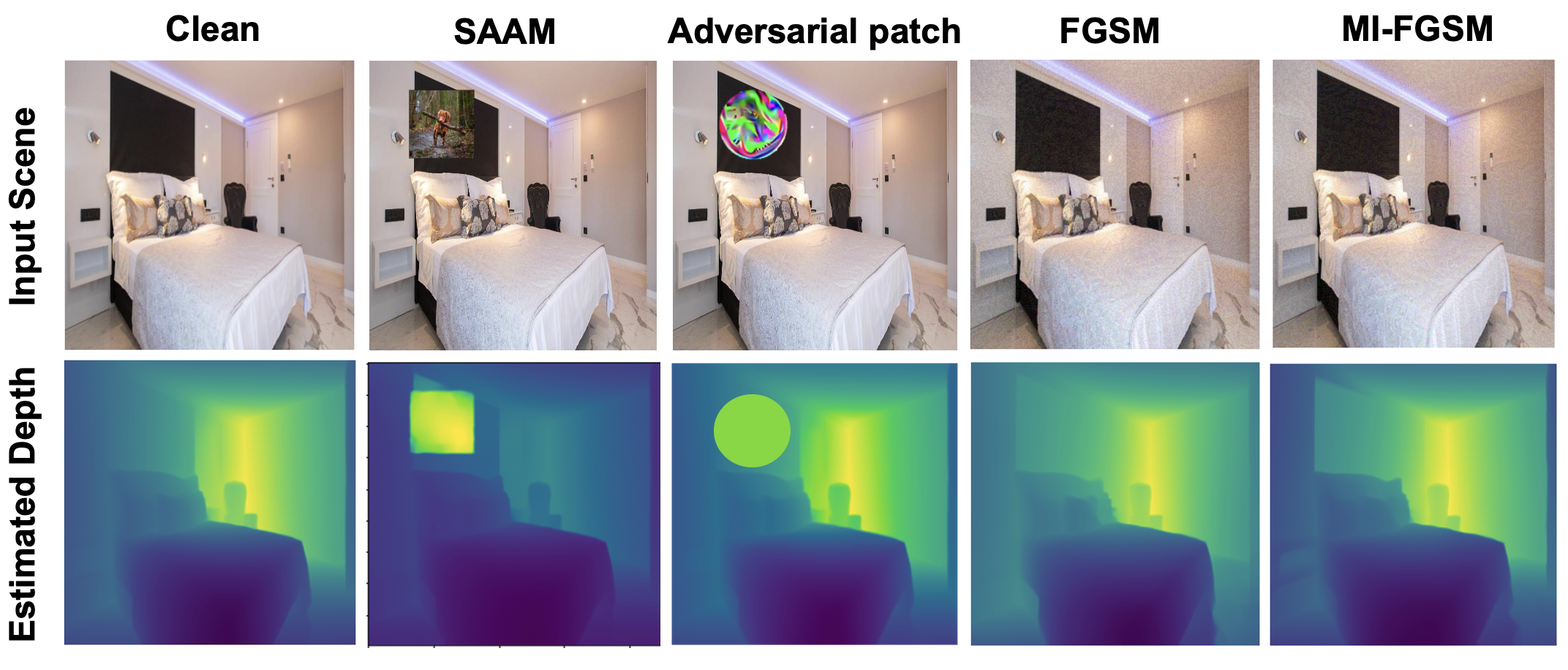}
    \caption{SAAM vs Adversarial patch \cite{Yamanaka_Access}, FGSM \cite{fgsm} and MI-FGSM \cite{dong2018boosting}: A comparison of the four methods in terms of their effectiveness and impact on model performance.}
    \label{fig:sota}
\end{figure*}
\subsection{SAAM vs Baseline attacks} 

\noindent \textbf{FGSM} \cite{fgsm} is a single-step, gradient-based, attack. An adversarial example is generated by performing a one step gradient update along the direction of the sign of gradient at each pixel as follows:

 \begin{equation}
     x^{adv} = x - \epsilon \cdot sign (\nabla_{x}J(x,y))
 \end{equation}
Where $\nabla J()$ computes the gradient of the loss function $J$ and $\theta$ is the set of model parameters. The $sign()$ denotes the sign function and $\epsilon$ is the perturbation magnitude. 

\noindent \textbf{MI-FGSM} \cite{dong2018boosting} introduced a momentum term to stabilize the update direction during the iteration.
 \begin{equation}
     g_{t+1} = \mu \cdot g_{t} + \frac{\nabla_{x}J(x_{t+1}^{adv},y)}{\parallel \nabla_{x}J(x_{t+1}^{adv},y) \parallel_1}
 \end{equation}
 \begin{equation}
     x_{t+1}^{adv} = x_{t}^{adv} - \alpha \cdot sign (g_{t+1})
 \end{equation}

We conducted a comparative analysis between our technique and FGSM and MI-FGSM. Our experiments involved running FGSM with a noise magnitude of epsilon = 8/255. For MI-FGSM, we used epsilon = 8/255, alpha = 2/255, executed 10 steps, and applied a decay factor of 1.0. 
As demonstrated in Table 1, our approach yields a depth estimation error of 0.58, whereas FGSM and MI-FGSM exhibit errors of 0.12 and 0.078, respectively. 
Our technique also achieves an impressive ratio of the affected region, registering at 0.99. In contrast, FGSM and MI-FGSM only manage to attain a modest ratio of 0.23 and 0.16, respectively.
 \begin{table}[!ht]
  \centering
  \caption{SAAM vs Baseline attacks: FGSM and MI-FGSM.}
  \label{vs}
  \begin{tabular}{|l|c|c|c|}
    \hline
    \textbf{Metric} & \textbf{SAAM}  &\textbf{FGSM \cite{fgsm}} &\textbf{MI-FGSM \cite{dong2018boosting}}\\
    \hline
      \textbf{$E_{d}$}  &  0.58    &   0.12   & 0.078 \\
      \hline
     \textbf{$R_{a}$}  &  0.99    &   0.23   &  0.16\\
  \hline
\end{tabular}
\end{table}

\subsection{SAAM vs Patch-based attacks}
 \begin{figure*}[!htp]
     \centering
     \includegraphics[width=0.65\textwidth]{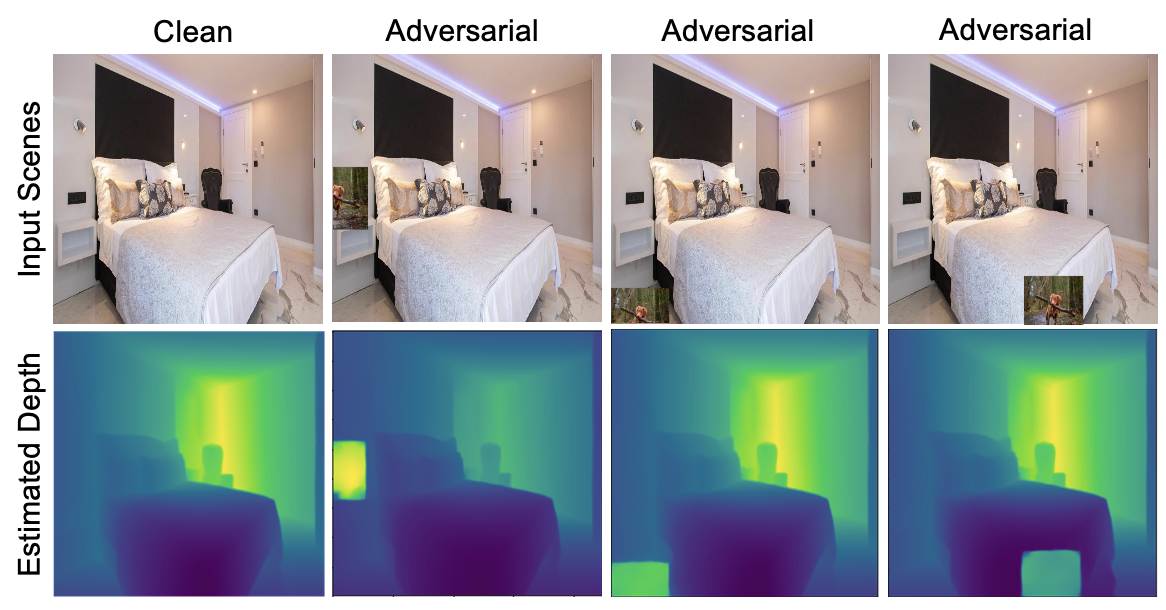}
     \caption{Impact of Occlusion on Patch effectiveness.}
     \label{fig:occlusion}
 \end{figure*}
In addition to its stealthiness, our adversarial patch also demonstrated superior performance in deceiving the monocular depth estimation model. Through careful design and optimization, the patch was able to effectively exploit the vulnerabilities of the model, leading to significantly altered and inaccurate depth predictions. Compared to other attack methods, our patch exhibited a higher success rate in fooling the depth estimation model, highlighting its effectiveness in undermining the model's performance. 
In fact, For a patch size of 5\%, SAAM achieves a substantial 58\% depth error, while the Adversarial patch method achieves a lower 40\% depth error.

 \begin{table}[!ht]
  \centering
  \caption{SAAM vs Patch-based attacks.}
  \label{vs}
  \begin{tabular}{|l|c|c|}
    \hline
    \textbf{Metric} & \textbf{SAAM}  &\textbf{Adversarial patch \cite{Yamanaka_Access}} \\
    \hline
      \textbf{$E_{d}$}  &  0.58    &   0.4     \\
      \hline
     \textbf{$R_{a}$}  &  0.99    &    0.98    \\
  \hline
\end{tabular}
\end{table}

\subsection{SAAM vs Random patch}

While adversarial patches are purposefully designed to exploit model vulnerabilities and achieve specific objectives, random patches are not designed with any specific objective in mind. They are typically generated by randomly selecting or generating patterns or images, without considering their impact on the model's output. Random patches lack the intentional manipulation and optimization seen in adversarial patches, and therefore do not possess the same level of effectiveness in influencing the model's decision-making process.
In Figure \ref{fig:random}, (1) \& (2) illustrates how placing random noise in the scene doesn't have an impact on the depth prediction. However, for (3) \& (4), SAAM is effective in manipulating the predicted depth even under perspective changes and rotations. 
Moreover, as indicated in Table \ref{vsrandom}, when quantitatively assessing the performance of our patch SAAM in comparison to a random patch, a notable disparity emerges. The random patch demonstrates minimal influence on the model's predicted depth compared to SAAM. In fact, our patch implementation achieves a depth estimation error ($E_{d}$) of $0.58$, whereas the random patch yields a mere $0.01$ error, with $R_{a} = 0$, signifying the ratio of pixels whose depth values have altered by more than $0.1$. Notably, SAAM achieves an impressive $R_{a}$ value of $0.99$, indicating that all pixels undergo a change surpassing $0.1$ in depth value.

Regardless of rotation, perspective changes, resizing, or any other applied transformation, our patch retains its effectiveness in manipulating the model's predictions. This robustness enables the patch to maintain its adversarial impact under different real-world conditions and ensures that it can be successfully deployed in a wide range of scenarios.

 \begin{table}[!ht]
  \centering
  \caption{SAAM vs Random patch.}
  \label{vsrandom}
  \begin{tabular}{|l|c|c|}
    \hline
    \textbf{Metric} & \textbf{SAAM}  &\textbf{Random patch} \\
    \hline
      \textbf{$E_{d}$}  &  0.58    &   0.01     \\
      \hline
     \textbf{$R_{a}$}  &  0.99    &    0    \\
  \hline
\end{tabular}
\end{table}


\section{Discussion}
\label{sec:discussion}
\subsection{Impact of Occlusion on Patch effectiveness}
Despite only a portion of the patch being visible in the captured image, its presence still had a significant impact on the depth prediction, leading to an inaccurate estimation. This observation highlights the vulnerability of the depth estimation model to adversarial perturbations, even when only a fraction of the perturbation is visible.

The effectiveness of the patch in influencing the depth prediction can be attributed to the model's reliance on local image features and its susceptibility to small perturbations in those features. The adversarial patch, carefully designed to exploit these vulnerabilities, can disrupt the model's perception of depth by introducing misleading cues or altering the local features relevant to depth estimation.

\subsection{SAAM robustness against Input Transformation-based Defenses}
We thoroughly evaluate the capability of our technique to withstand diverse sets of transformations, we use three commonly used defense approaches that perform input transformations without re-training the victim models to assess the robustness of our patch against these defenses. We use JPEG compression \cite{jpeg}, add Gaussian noise\cite{gaussian}, Median blurring \cite{median}. 

The outcomes of our experiments, as showcased in Tables \ref{performance_JPEG}, \ref{performance_MEDIAN}, and \ref{performance_GAUSSIAN}, substantiate the superiority of our proposed technique over the adversarial patch by Yamanaka et al. \cite{Yamanaka_Access}. Our method consistently yields a higher attack success rate across the diverse transformations considered. This serves as compelling evidence of the resilience and effectiveness of our approach when facing real-world distortions, affirming its potential for practical application in adversarial attacks on the targeted system.

\begin{table}[!htp]
  \centering
  \caption{Mean depth error when applying JPEG compression.}
  \label{performance_JPEG}
  \begin{tabular}{|c|c|c|c|c|}
    \hline
    \textbf{Parameters} &  \textbf{90} & \textbf{70} & \textbf{50}  & \textbf{30} \\
    \hline
      $E_{d}$  & 0.25 & 0.23  & 0.2  & 0.17  \\
  \hline
\end{tabular}
\end{table}
 
\begin{table}[!htp]
  \centering
  \caption{Mean depth error when applying median blur.}
  \label{performance_MEDIAN}
  \begin{tabular}{|c|c|c|c|c|}
    \hline
    \textbf{Parameters} &  \textbf{5} & \textbf{10} & \textbf{15}  & \textbf{20} \\
    \hline
      $E_{d}$  & 0.18  & 0.16  & 0.15  & 0.14  \\
  \hline
\end{tabular}
\end{table}

\begin{table}[!htp]
  \centering
  \caption{Mean depth error when applying Gaussian noise.}
  \label{performance_GAUSSIAN}
  \begin{tabular}{|c|c|c|c|c|}
    \hline
    \textbf{Parameters} &  \textbf{0.01} & \textbf{0.02} & \textbf{0.05}  & \textbf{0.1} \\
    \hline
      $E_{d}$  & 0.25  & 0.23  &  0.21 & 0.2 \\
  \hline
\end{tabular}
\end{table}

\subsection{Future Work}

In our research, we have introduced the Stealthy Adversarial Attack on Monocular Depth Estimation (SAAM) technique, a significant advancement in the field of adversarial attacks on monocular depth estimation (MDE). Yet, there are promising directions for future exploration:
\begin{itemize}

\item  Defense Mechanism Development: Future research can be dedicated to the creation and evaluation of robust defense strategies capable of effectively countering stealthy adversarial patches like SAAM, ensuring the integrity and reliability of MDE systems.

\item  Multi-Objective Attacks: An intriguing avenue is the investigation of adversarial patches designed to simultaneously impact depth estimation and other computer vision tasks, such as object detection or semantic segmentation. Multi-objective attacks present complex challenges and have profound implications for overall system reliability.

\item  Cross-Modality Threats: Research can delve into cross-modality adversarial attacks, where patches initially designed for MDE are extended to interfere with other sensors like lidar or radar. This expansion of the attack vector poses a more comprehensive security threat, particularly in the context of sensor fusion systems.
\end{itemize}
These research directions promise to further our understanding of adversarial attacks in MDE and contribute to the development of robust countermeasures to safeguard critical applications.
\section{Conclusion}
\label{sec:conclusion}

In this paper, we introduce a novel patch-based adversarial attack named SAAM, specifically designed to compromise Monocular Depth Estimation (MDE)-based vision systems. SAAM is a carefully crafted patch that can completely conceal objects or manipulate their perceived depth within a scene.
The experimental results confirm that the SAAM patch can successfully compromise MDE-based vision systems, highlighting the vulnerability of such systems to physical adversarial attacks. This research sheds light on the potential risks posed by adversarial attacks in real-world applications where MDE plays a crucial role, such as autonomous vehicles, robotics, and augmented reality.
In fact, our patch achieves almost 60\% mean depth estimation error, with almost 100$\%$ of the target region being affected. 
Our proposed method is designed to be applicable to the real world and practical scenarios. The generated adversarial patch is optimized to be effective in real-world settings and is designed to deceive depth estimation systems when observed through cameras or sensors. Once the patch is generated, it can be easily printed and used in the physical world without the need for any specialized equipment or complex setup. 
By demonstrating the susceptibility of depth estimation systems to such adversarial attacks, our research emphasizes the need for robust defense mechanisms to enhance the reliability and security of depth-based computer vision applications, including those in autonomous vehicles, surveillance, and augmented reality.

\section*{Acknowledgment}

This research was partially funded by Technology Innovation Institute (TII) under the "CASTLE: Cross-Layer Security for Machine Learning Systems IoT" project.

\bibliographystyle{unsrt}
\bibliography{bib.bib}
\begin{IEEEbiography}[{\includegraphics[width=1in,height=1.25in,clip,keepaspectratio]{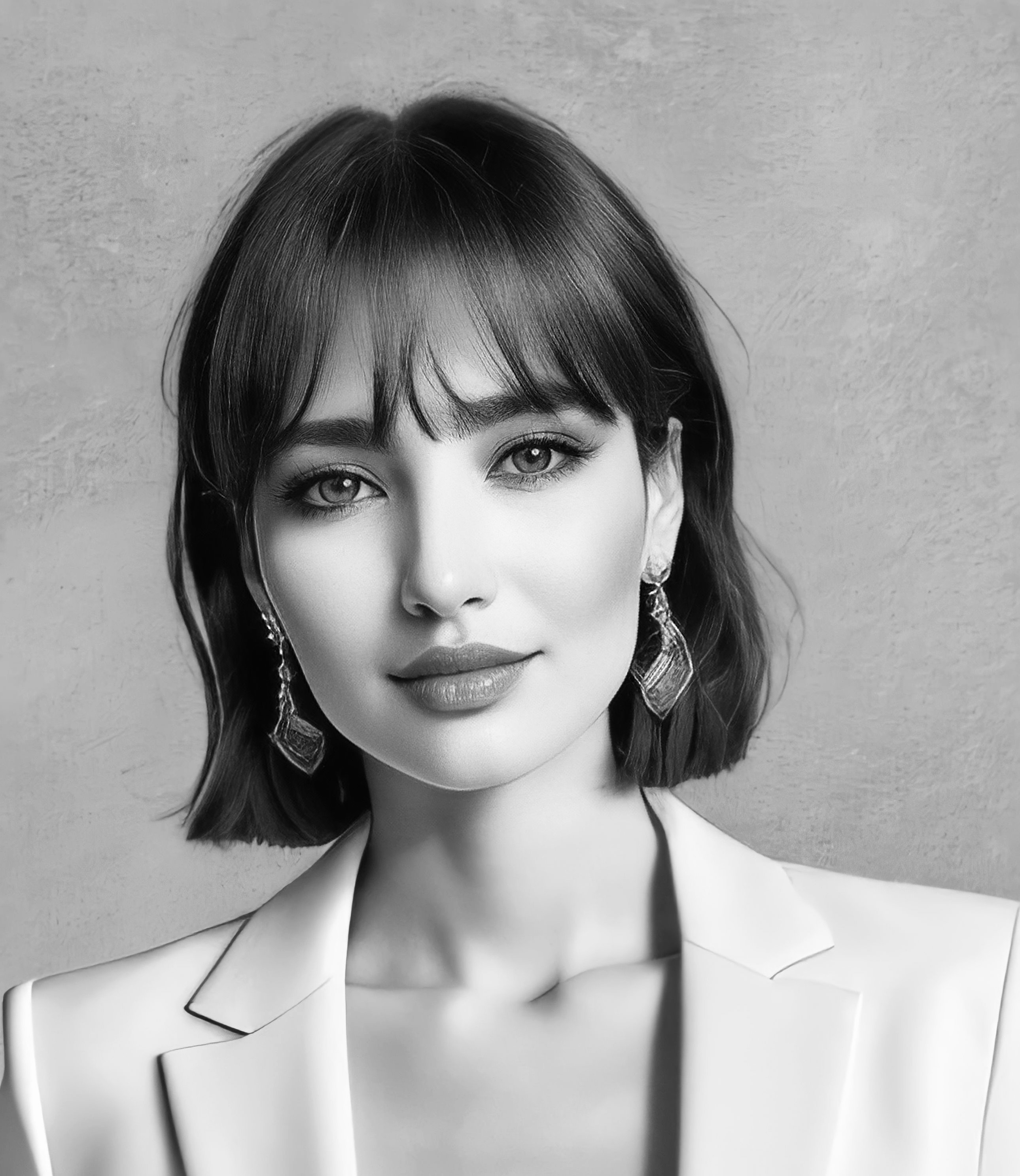}}]{Amira Guesmi} received the Engineer degree in Computer Science \& Electrical Engineering, from National School of Engineers of Sfax, Tunisia, in 2016, and the Ph.D. degree in Computer Systems Engineering from Polytechnic University Hauts-De-France, France, and the National School of Engineers of Sfax, Tunisia, in 2021. Afterwards, she worked as a Postdoctoral researcher at IEMN-DOAE Laboratory (CNRS-8520), Université Polytechnique Hauts-de-France. 
Dr Guesmi is currently working as a research group leader in New York University (NYU) Abu Dhabi, UAE. Her research interests include AI safety, machine learning security and Privacy, lifelong learning, approximate computing, and energy-efficient design.

\end{IEEEbiography}

\begin{IEEEbiography}[{\includegraphics[width=1in,height=1.25in,clip,keepaspectratio]{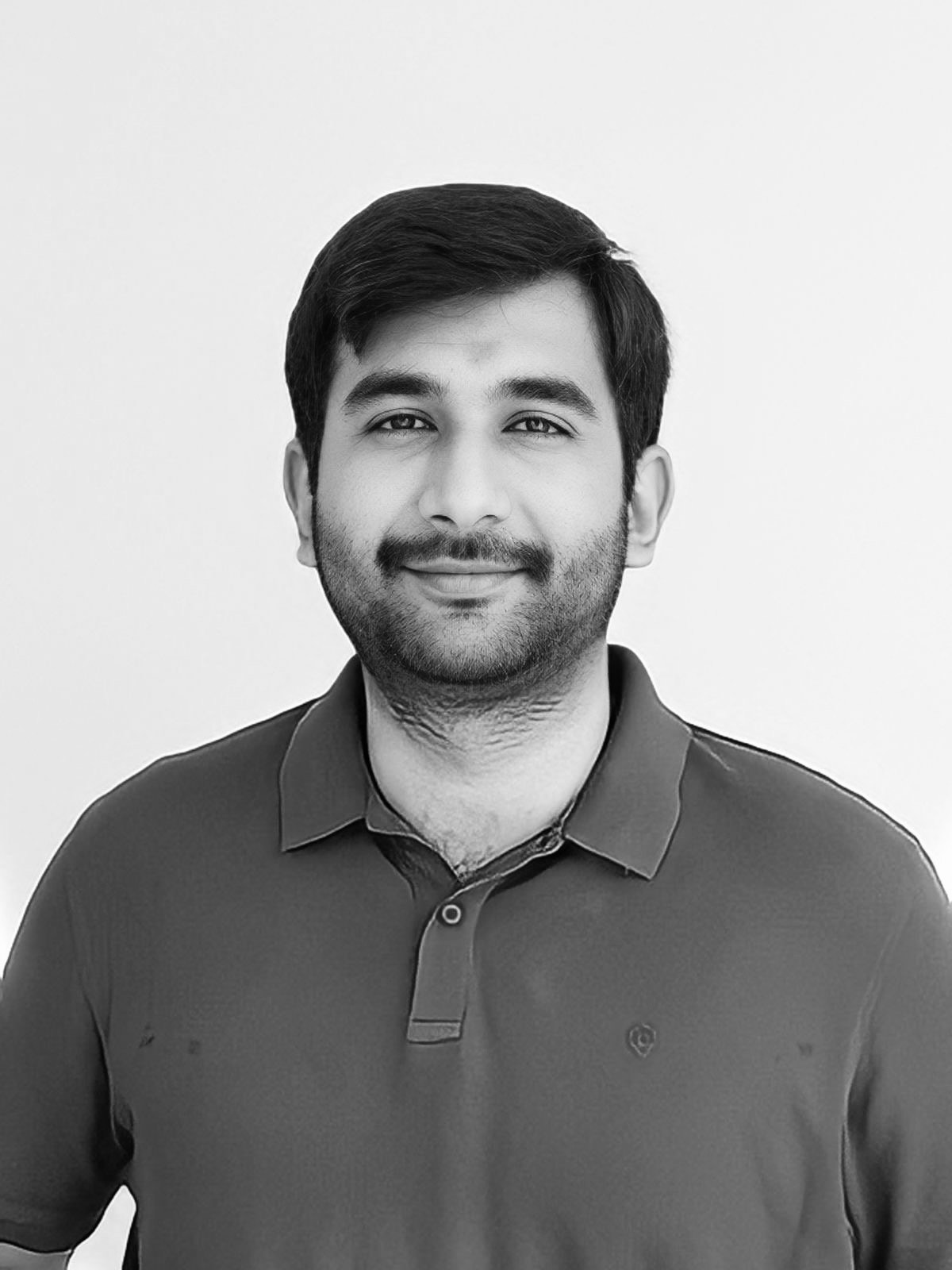}}]{Muhammad Abdullah Hanif} received the B.Sc. degree in Electronic Engineering from the Ghulam Ishaq Khan Institute of Engineering Sciences and Technology (GIKI), Pakistan, and the M.Sc. degree in Electrical Engineering with specialization in digital systems and signal processing from the School of Electrical Engineering and Computer Science, National University of Sciences and Technology (NUST), Islamabad, Pakistan. Then, he studied for his Ph.D. in Computer Engineering at Vienna University of Technology (TU Wien), Austria. He is currently working as a research group leader in New York University (NYU) Abu Dhabi, UAE. His research interests include brain-inspired computing, machine learning, approximate computing, computer architecture, energy-efficient design, robust computing, system-on-chip design, and emerging technologies. 
\end{IEEEbiography}

\begin{IEEEbiography}[{\includegraphics[width=1in,height=1.25in,clip,keepaspectratio]{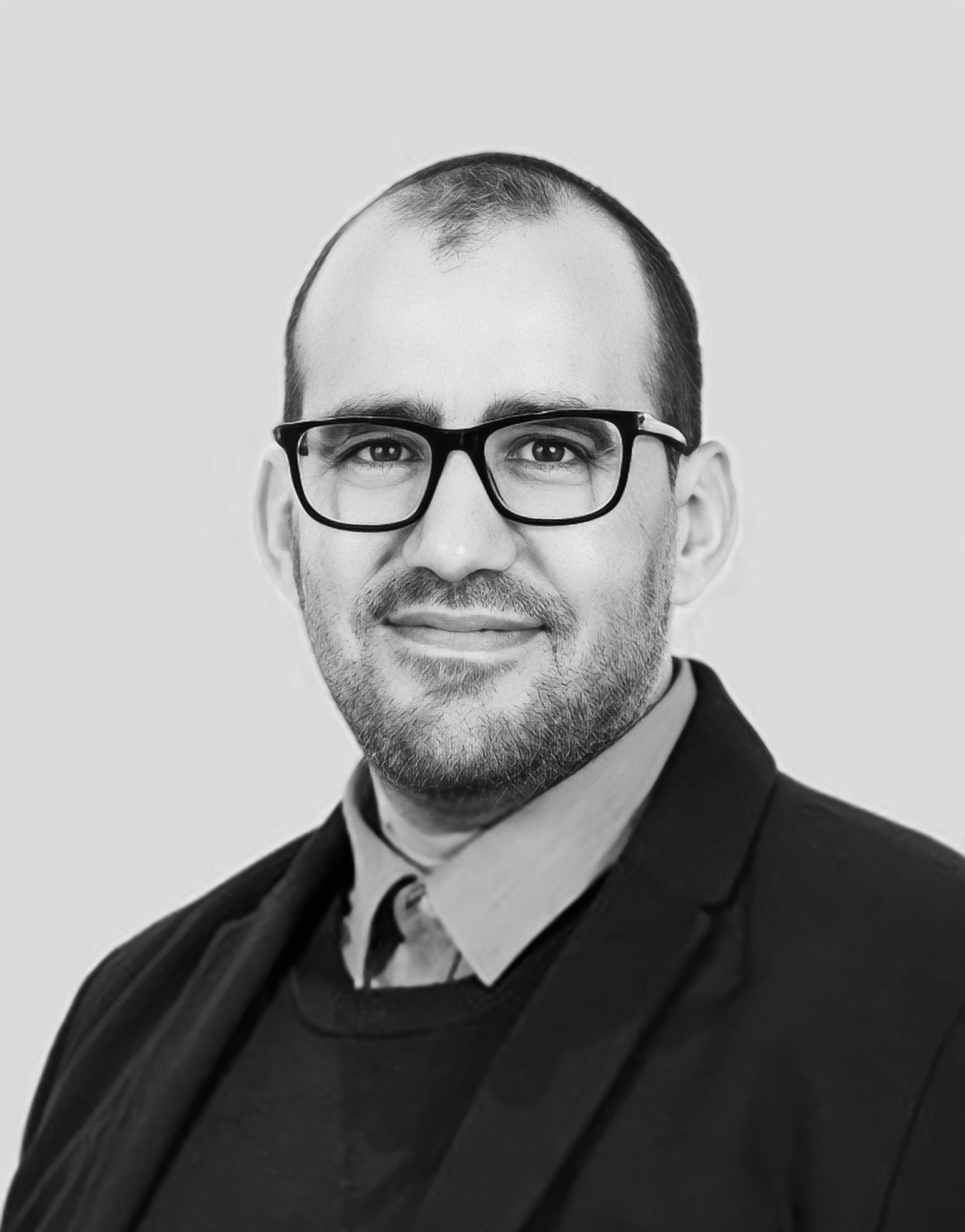}}]{Bassem Ouni} Dr./Eng. Bassem Ouni is currently a Lead Researcher at Technology Innovation Institute, Abu Dhabi,
United Arab Emirates. He received his Ph.D. degree in Computer Science from the University of Nice-Sophia Antipolis, Nice, France, in July 2013. Between October 2018 and January 2022, he held a Lead Researcher position in the French Atomic Energy Commission (CEA) within the LIST Institute, Paris, France, and an Associate Professor/Lecturer position at the University of Paris Saclay and ESME Sudria Engineering school, Paris, France. Prior to that, he worked as a Lead Researcher between 2017 and 2018 within the department of Electronics and Computer Science, University of Southampton, Southampton, United Kingdom. Before that, he occupied the position of a Research Scientist, between 2015 and 2016, at the Institute of Technology in Aeronautics, Space and Embedded Systems (IRT-AESE) located in Toulouse, France. From September 2013 to the end of 2014, he held a post-doctoral fellow position in Eurecom, Telecom ParisTech institute, Sophia Antipolis, France. Furthermore, he worked between 2009 and 2013 as a lecturer at the University of Nice Sophia Antipolis (Polytech Nice Engineering School and Faculty of
Sciences of Nice). Also, he was managing several industrial collaborations with ARM, Airbus Group Innovation, Rolls Royce, Thales Group, Continental, Actia Automotive Group, etc. He co-authored many publications (Book Chapters, Journals, and international conferences.). He is an IEEE senior member.
\end{IEEEbiography}

\begin{IEEEbiography}[{\includegraphics[width=1in,height=1.25in,clip,keepaspectratio]{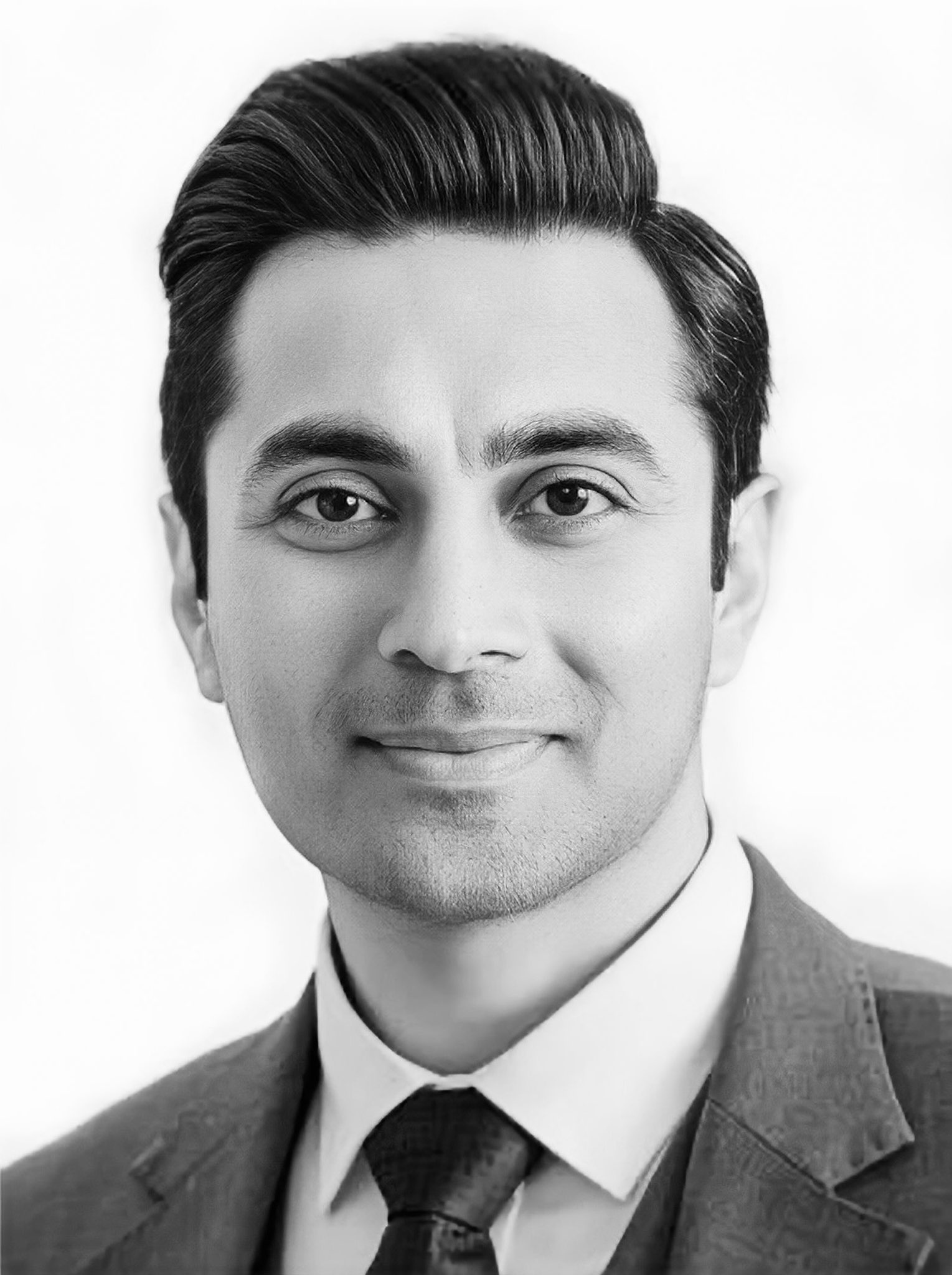}}]{Muhammad Shafique} (M’11 - SM’16) received the Ph.D. degree in computer science from the Karlsruhe Institute of Technology (KIT), Germany, in 2011. Afterwards, he established and led a highly recognized research group at KIT for several years as well as conducted impactful collaborative R\&D activities across the globe. In Oct.2016, he joined the Institute of Computer Engineering at the Faculty of Informatics, Technische Universität Wien (TU Wien), Vienna, Austria as a Full Professor of Computer Architecture and Robust, Energy-Efficient Technologies. Since Sep.2020, Dr. Shafique is with the New York University (NYU), where he is currently a Full Professor and the director of eBrain Lab at the NYU-Abu Dhabi in UAE, and a Global Network Professor at the Tandon School of Engineering, NYU-New York City in USA. He is also a Co-PI/Investigator in multiple NYUAD Centers, including Center of Artificial Intelligence and Robotics (CAIR), Center of Cyber Security (CCS), Center for InTeractIng urban nEtworkS (CITIES), and Center for Quantum and Topological Systems (CQTS).
His research interests are in AI \& machine learning hardware and system-level design, brain-inspired computing, machine learning security and privacy, quantum machine learning, cognitive autonomous systems, wearable healthcare, energy-efficient systems, robust computing, hardware security, emerging technologies, FPGAs, MPSoCs, and embedded systems. His research has a special focus on cross-layer analysis, modeling, design, and optimization of computing and memory systems. The researched technologies and tools are deployed in application use cases from Internet-of-Things (IoT), Smart Cyber-Physical Systems (CPS), and ICT for Development (ICT4D) domains. Dr. Shafique has given several Keynotes, Invited Talks, and Tutorials, as well as organized many special sessions at premier venues. He has served as the PC Chair, General Chair, Track Chair, and PC member for several prestigious IEEE/ACM conferences. Dr. Shafique holds one U.S. patent, and has (co-)authored 6 Books, 10+ Book Chapters, 350+ papers in premier journals and conferences, and 100+ archive articles. He received the 2015 ACM/SIGDA Outstanding New Faculty Award, the AI 2000 Chip Technology Most Influential Scholar Award in 2020 and 2022, the 2015 ACM/SIGDA Outstanding New Faculty Award, the AI 2000 Chip Technology Most Influential Scholar Award in 2020, 2022, and 2023, the ASPIRE AARE Research Excellence Award in 2021, six gold medals, and several best paper awards and nominations at prestigious conferences. He is a senior member of the IEEE and IEEE Signal Processing Society (SPS), and a member of the ACM, SIGARCH, SIGDA, SIGBED, and HIPEAC.

\end{IEEEbiography}

\EOD

\end{document}